# Generative Discovery of Partial Differential Equations by Learning from Math Handbooks

Hao Xu[1,2], Yuntian Chen[1,3,*], Rui Cao[4,5], Tianning Tang[6,7], Mengge Du[8], Jian Li[3], Adrian H. Callaghan[5], and Dongxiao Zhang[1,9,*]

[1] Zhejiang Key Laboratory of Industrial Intelligence and Digital Twin, Eastern Institute of Technology, Ningbo, Zhejiang 315200, P. R. China

[2] Department of Electrical Engineering, Tsinghua University, Beijing 100084, P. R. China

[3] Ningbo Institute of Digital Twin, Eastern Institute of Technology, Ningbo, Zhejiang 315200, P. R. China

[4] College of Oceanic and Atmospheric Sciences, Ocean University of China, Qingdao 266100, P. R. China

[5] Department of Civil and Environmental Engineering, Imperial College London, London, SW7 2AZ, United Kingdom

[6] Department of Engineering Science, University of Oxford, Parks Road, Oxford, OX1 3PJ, United Kingdom

[7] Department of Mechanical and Aerospace Engineering, University of Manchester, Manchester, M13 9PL, United Kingdom

[8] College of Engineering, Peking University, Beijing 100871, P. R. China

[9] Institute for Advanced Study, Lingnan University, Tuen Mun, Hong Kong

[*] Corresponding authors

Email address: ychen@eitech.edu.cn (Y. Chen); dzhang@eitech.edu.cn (D. Zhang)

**Abstract**

Data-driven discovery of partial differential equations (PDEs) is a promising approach for uncovering the underlying laws governing complex systems. However, purely data-driven techniques face the dilemma of balancing search space with optimization efficiency. This study introduces a knowledge-guided approach that incorporates existing PDEs documented in a mathematical handbook to facilitate the discovery process. These PDEs are encoded as sentence-like structures composed of operators and basic terms, and used to train a generative model, called EqGPT, which enables the generation of free-form PDEs. A loop of "generation–evaluation–optimization" is constructed to autonomously identify the most suitable PDE. Experimental results demonstrate that this framework can recover a variety of PDE forms with high accuracy and computational efficiency, particularly in cases involving complex temporal derivatives or intricate spatial terms, which are often beyond the reach of conventional methods. The approach also exhibits generalizability to irregular spatial domains and higher dimensional settings. Notably, it succeeds in discovering a previously unreported PDE governing strongly nonlinear surface gravity waves propagating toward breaking, based on real-world experimental data, highlighting its applicability to practical scenarios and its potential to support scientific discovery.

**Introduction**

Partial differential equations (PDEs) are important tools for describing complex dynamic processes in nature and constitute one of the cornerstones of scientific research. For a long time, PDEs have been developed on the basis of first-principle derivation, which typically relies on manual effort. In recent years, with improvements in data accessibility and computational science, discovering PDEs directly from high-fidelity observations has become a promising way to identify potential governing laws in complex systems of diverse fields[1–3]. The discovered PDEs possess better interpretability and generalizability than black-box models, and are able to reveal undisclosed mechanisms and insights, such as conservation laws and symmetries[4].

The identification of underlying PDEs from data typically involves two key aspects: determining the relevant terms; and capturing the relationships among them. In the literature, sparse regression techniques are introduced to address this by constructing a candidate library based on *a priori* selection of potential terms, and subsequently identifying a parsimonious structure through $L_0$ or $L_1$ normalization[5,6]. Despite its computational efficiency and straightforward implementation, sparse regression faces limitations in practice since it is often infeasible to enumerate all possible candidate terms, given the diversity of nonlinear interactions and composite expressions in derivative terms. When more candidate terms are included, the iterative elimination of terms with small coefficients may fail to converge to a correct equation. While weak-form variants have been proposed to improve stability[7–9], these are often constrained to polynomial or high-order derivatives, and struggle to handle more intricate terms, such as $u_{xxt}$ or $xu_t$. In essence, the effectiveness of sparse regression is closely tied to the completeness and design of the candidate library—yet constructing such a library remains a challenge, particularly for systems with complex dynamics. In response to these limitations, some symbolic regression-based approaches, including genetic algorithms[10,11] and tree-based search strategies[12–14], are proposed to discover PDEs in a free-form manner. While this alleviates the dependence on predefined libraries and substantially expands the search space, it also incurs optimization challenges, particularly under high-noise conditions or in multivariable systems.

Recent advances in generative models have introduced new opportunities for data-driven equation discovery[15,16]. Current efforts can be broadly categorized into three approaches. The first is "prompt-based discovery," which leverages general-purpose large language models (LLMs), such as GPT-4 or Gemini[17–19]. They can infer potential candidates based on prompt templates that encode problem context and inquires. However, such approaches rely heavily on human involvement for prompt design and output evaluation, and they typically lack the ability to directly process observational data, resulting in high cost and limited efficiency. The second is "data-to-sequence discovery," in which generative models are trained to directly map from data to equation structures[20,21]. While inference is fast, this requires large-scale paired datasets of PDEs and their solutions for training. Given the high computational cost of PDE simulation, such models are currently limited in scope and accuracy, and often restricted to a narrow range of canonical equations. The third is "sequence optimization discovery," in which generative models produce candidate equations that are subsequently optimized using reinforcement learning, such as risk-seeking policy gradient[22–25]. However, this approach inherits the difficulties of symbolic regression. In PDE discovery, the primary source of complexity arises not from mathematical nesting, but from intricate combinations of differential operators. Prefix tree-based optimization strategies often introduce excessive search complexity and spend computational resources on physically implausible terms, which results in poor robustness and efficiency, particularly in noisy, high-dimensional scenarios.

Taken together, these challenges reveal a common limitation across both traditional sparse regression and emerging LLM-based approaches: a purely data-driven search process often faces a trade-off between search space size and optimization efficiency. While recent methods have improved optimization techniques, they still operate largely independently of domain-specific prior knowledge. In this work, we propose a different strategy. Rather than focusing solely on optimization, we explore how existing mathematical and scientific knowledge can be embedded into the equation discovery process. Our central idea is inspired by the notion that "stones from other hills may polish one's own jade"—namely, that general knowledge of PDE structures, summarized from established scientific literature, can guide the discovery of new equations in specific contexts.

Specifically, we present a framework that integrates knowledge-guided and data-driven strategies for PDE discovery (Fig. 1). Two techniques are introduced, including generative representation of equation (GRE) and scientifically augmented training (SAT). In the GRE, a structured equation encoding scheme is proposed, wherein equations are parsed into vocabularies composed of operators (e.g., arithmetic symbols) and fundamental physical terms. These units are combined to form sequence representations of free-form equations. In the SAT, a generative model, called EqGPT, is trained to learn co-occurrence patterns among PDE terms from a dataset of PDEs collected from a mathematical handbook[26]. This enables the autonomous generation of free-form candidate equations while implicitly filtering out physically implausible expressions, thereby enhancing the efficiency and relevance of the search process and avoiding brute-force enumeration. Through a knowledge-guided loop of "generation–evaluation–optimization", PDEs that are both consistent with observed data and aligned with domain knowledge can be identified autonomously.

Proof-of-concept experiments validate the framework's ability to rediscover canonical PDEs in irregular computational domains and to recover multi-variable spatiotemporal equations in high-dimensional settings. Comparative analysis with existing baseline approaches demonstrates its improved accuracy and robustness. Furthermore, we present a practical case study, in which the proposed approach successfully uncovers a previously unreported governing equation for highly nonlinear surface gravity waves propagating toward breaking from real-world experimental data. This example illustrates the proposed framework's practical utility and provides concrete evidence that it can contribute to scientific understanding by revealing new relationships.

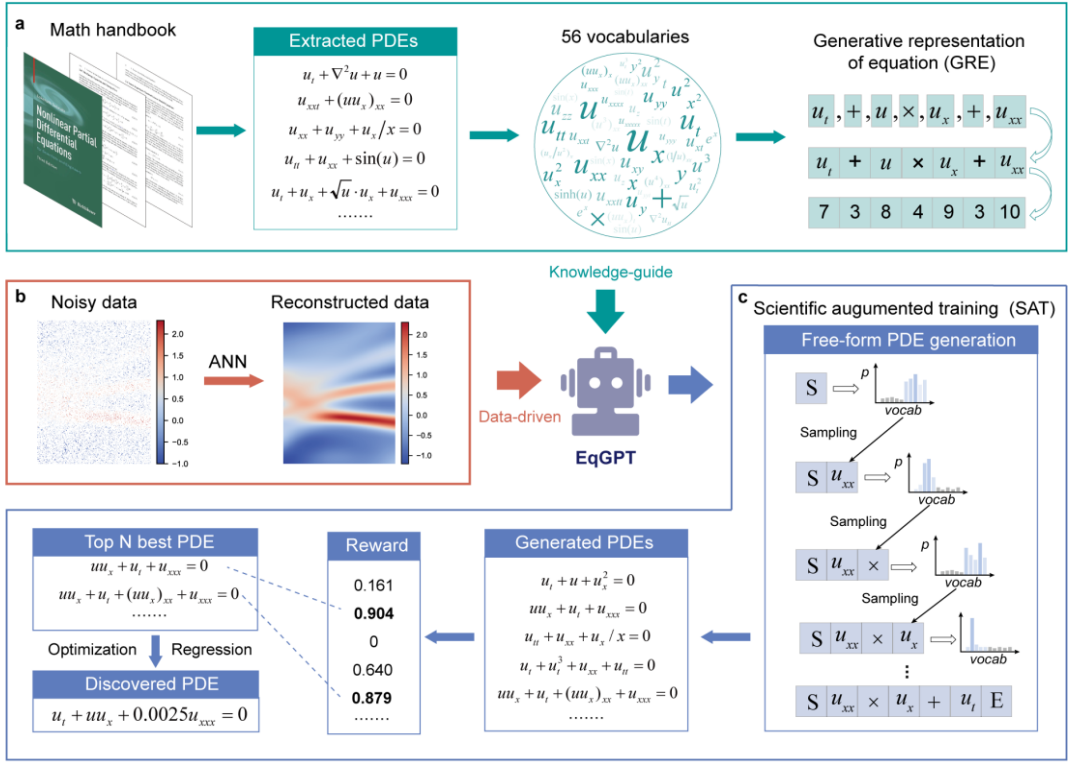

**Fig. 1. Overview of the proposed knowledge-guided generative framework for PDE discovery. (a)** The construction of knowledge guidance, including extracting existing PDEs from a math handbook, which is tokenized into 56 vocabularies (basic terms) and converted to sentences through generative representation of equations

(GRE). **(b)** The data-driven workflow, in which sparse and noisy data are reconstructed by an artificial neural network (ANN). **(c)** The loop of "generation–evaluation–optimization", including knowledge-guided generation of free-form PDEs, data-driven evaluation, and model fine-tuning.

## Results

### The proposed knowledge-guided generative framework for PDE discovery

The research framework, as depicted in Fig. 1, consists of three key workflows, including knowledge-guided pathway, data-driven pathway, and combined optimization approach. In the knowledge-guided pathway, the scientific knowledge of existing PDEs is incorporated by generative representation of equations (GRE). Therefore, a dataset of 221 PDE structures is extracted from the scientific handbook *Nonlinear Partial Differential Equations*[26], which contains canonical PDEs in science and engineering. Through 56 vocabularies obtained by tokenizing PDEs into operators and basic terms, PDEs can be converted into sentences (Fig. 1a). Considering that the position of the terms in the equation does not affect the structure or meaning of the equation itself, we performed data augmentation by swapping the positions of the terms in the equations, ultimately generating 7,072 sentences. The encoded PDE structures are utilized to train the generative model, called EqGPT. Here, the architecture of the GPT- model is adopted[27]. Notably, EqGPT functions as both a knowledge learner and an equation generator, with its architecture extending beyond the constraints of GPT-2. The focus of this work is not on introducing a new neural architecture, but rather on illustrating how domain-specific knowledge, especially PDE structures, can be effectively embedded through pretraining to facilitate the equation discovery process. In the training process, the feature distribution of the PDE dataset and the co-occurrence probability of vocabularies are learned by predicting each token from the preceding token(s). The generative model is trained for 100 epochs in this work. The model can be subsequently utilized to generate new PDE structures. In the data-driven pathway, considering that observation data may be sparse and noisy, a fully connected artificial neural network (ANN) is trained with the observation data to function as a surrogate model, which is essentially a data generator that can smooth the data through interpolation. The surrogate model can not only generate abundant predictions on grids (meta-data), but also calculate derivatives through automatic differentiation in back propagation[28]. This facilitates the evaluation of discovered PDEs and provides robustness to high levels of noise.

For the optimization of the most suitable PDEs, the knowledge-guided and data-driven pathways are combined to form a closed loop of "generation–evaluation–optimization". In the generation process, the start token (S) is first input into the model, and the next token is sampled from the predicted probability distribution. The model repeatedly samples tokens from the learned distribution until the end token is sampled, indicating the completion of a new PDE structure (Fig. 1c). Here, the probabilistic sampling and random sampling are combined to guarantee the variety of generative PDEs. Benefitted from the proposed GRE technique for PDEs, EqGPT can generate diversified free-form PDEs. A reward function is defined to evaluate the generated

PDEs:

$$\text{Reward} = \alpha \cdot R^2 = \left(1 - \alpha_0 \log_{10} N_{term}\right) \cdot \left(1 - \frac{\sum (\theta_{LHS} - \bar{\bar{\xi}} \cdot \theta_{RHS})^2}{\sum (\theta_{LHS} - \bar{\theta}_{LHS})^2}\right). \qquad (1)$$

where the reward is composed of two components: the penalty term $\alpha$ and the determination coefficient $R^2$. The former is defined to measure the sparsity, which penalizes the number of terms in the structure to obtain a parsimonious PDE, and the latter is employed to measure the consistency with observation data through the $R^2$ of the regression. In the regression, $\theta_{LHS}$ is the negative of the first term of the structure; $\theta_{RHS}$ are the remaining terms; and $\bar{\bar{\xi}}$ is the coefficient obtained from the least squares regression. $\bar{\theta}_{LHS}$ is the mean value of $\theta_{LHS}$. Overall, the higher is the reward, the better is the generated PDE.

After the reward is calculated, the top 10 equations with the highest rewards are selected to fine-tune the EqGPT model (Fig. 1c). In other words, they are utilized as the training data to further adjust the model parameters with a smaller learning rate, $10^{-5}$. The fine-tuning epoch is 5. Since the number of training data for fine-tuning is tiny (only 10), the fine-tuning process is fast (only several seconds). The fine-tuned model then generates new PDE structures and updates the records of the top 10 structures, which constitute an optimization cycle. When the optimization cycle reaches the maximum number of epochs (5 in this work), the structure with the highest reward is the optimal PDE. On this basis, the coefficients of the best PDE structure are regressed and optimized to obtain the final PDE form. Additional details about the framework can be found in the Materials and Methods.

**Discovery of canonical PDEs with sparse and noisy data**

For proof-of-concept, 8 canonical PDEs from different scientific fields are employed to examine the performance of the proposed framework with sparse and noisy data. The true equation form, discovered equation form, noise robustness, and data robustness for discovering each PDE are illustrated in Fig. 2. Here, the discovered equation represents the outcome of the framework applied to 10,000 data points. The noise is Gaussian noise, which is defined as:

$$\tilde{u} = \gamma \cdot std(u) \cdot N(0,1) + u \qquad (2)$$

where $u$ represents the clean data; $\tilde{u}$ represents the noisy data; $N(0,1)$ represents the standard normal distribution; and $\gamma$ represents the noise level. To assess the noise robustness, the noise level is systematically elevated by 25% increments until the identification of the correct PDE form becomes unattainable. To examine the data robustness, the number of training data is decreased progressively as 10000, 5000, 2500, 1000, 500, and 100 until the correct PDE form fails to be discovered. If the equation form is correctly identified, the coefficient error is employed to measure the accuracy of the discovered PDE, which is defined as:

$$\varepsilon = \frac{1}{N_{term}} \sum_{i=1}^{N_{term}} \left| \frac{\xi_i - \xi_i^{true}}{\xi_i^{true}} \right|, \qquad (3)$$

where $N_{term}$ is the number of PDE terms; $\xi_i$ is the coefficient of the discovered PDE terms; and $\xi_i^{true}$ is the true PDE term. Notably, in the experiments, the target PDE is deleted from the dataset to ensure that the EqGPT model has never seen the target PDE, which guarantees the validity of the experiments. However, experiments have shown that the discovery of PDEs with and without target PDEs in the PDE dataset does not significantly influence the ultimate results (Supplementary Information S1.2).

Upon analysis of the results, we can derive several key insights. First, the proposed framework can discover correct PDEs accurately with more than 50% noise and fewer than 1,000 datapoints in most cases (Fig. 2a-f). This robustness is greater than that of existing mainstream PDE discovery algorithms[29–31], which confirms the general ability of our proposed framework to discover PDEs under sparse and noisy data; this is attributed to the surrogate model utilized in the proposed framework, since it can reconstruct the underlying process from sparse and noisy observations, which smooths the noise and generates sufficient meta-data on grids for PDE discovery. Additional experiments confirm the robustness of the proposed method under conditions of both data sparsity and high noise levels, as detailed in Supplementary Information S1.5. Moreover, the EqGPT model can generate proper structures, which simplifies the optimization process and facilitates the identification of the correct structures.

Notably, the proposed framework can discover PDEs with broader forms, such as uncommon derivatives and terms with fractions, which were previously difficult to discover. For example, the PDE in Fig. 2g contains a rarely seen derivative term $u_{xt}$, which was not discovered in preceding works. Without any prior knowledge, existing knowledge discovery algorithms have difficulty in handling such situations since they require predetermined terms for regression and usually assume that the left-hand side term is $u_t$ or $u_{tt}$[32,33]. In contrast, the current framework successfully identifies this PDE, as it learns the term $u_{xt}$ when training the generative model, although its origin could be traced to other equations, as well.

Fig. 3e provides another PDE that involves fraction terms, $u_x/x$. It is also difficult to discover by sparse regression or genetic evolution algorithms since they cannot generate free-form fraction terms. The proposed framework is able to handle this problem in a straightforward manner because of the operator "/" defined in the vocabularies. Furthermore, the proposed framework remains stable and accurate even with 100 data points for this complex situation, which confirms its ability to handle broader forms of PDE. The proposed framework can also discover PDEs in compound form, where the expanded form is also discovered as the suboptimal structure (Supplementary Information S1.4).

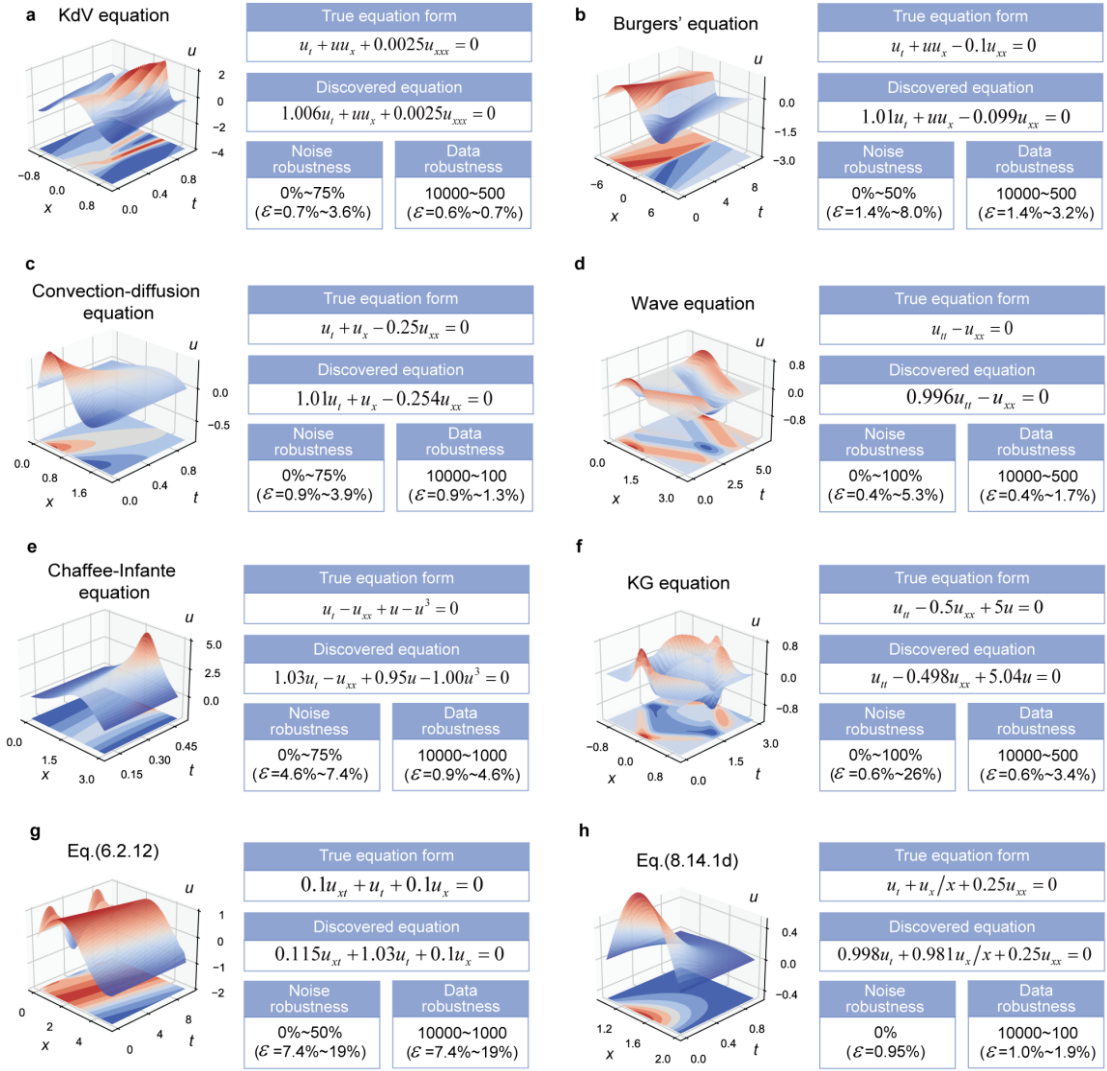

**a** KdV equation

| True equation form |
| --- |
| $u_t + uu_x + 0.0025u_{xxx} = 0$ |

| Discovered equation |
| --- |
| $1.006u_t + uu_x + 0.0025u_{xxx} = 0$ |

| Noise robustness | Data robustness |
| --- | --- |
| 0%~75% ($\varepsilon$=0.7%~3.6%) | 10000~500 ($\varepsilon$=0.6%~0.7%) |

**b** Burgers' equation

| True equation form |
| --- |
| $u_t + uu_x - 0.1u_{xx} = 0$ |

| Discovered equation |
| --- |
| $1.01u_t + uu_x - 0.099u_{xx} = 0$ |

| Noise robustness | Data robustness |
| --- | --- |
| 0%~50% ($\varepsilon$=1.4%~8.0%) | 10000~500 ($\varepsilon$=1.4%~3.2%) |

**c** Convection-diffusion equation

| True equation form |
| --- |
| $u_t + u_x - 0.25u_{xx} = 0$ |

| Discovered equation |
| --- |
| $1.01u_t + u_x - 0.254u_{xx} = 0$ |

| Noise robustness | Data robustness |
| --- | --- |
| 0%~75% ($\varepsilon$=0.9%~3.9%) | 10000~100 ($\varepsilon$=0.9%~1.3%) |

**d** Wave equation

| True equation form |
| --- |
| $u_{tt} - u_{xx} = 0$ |

| Discovered equation |
| --- |
| $0.996u_{tt} - u_{xx} = 0$ |

| Noise robustness | Data robustness |
| --- | --- |
| 0%~100% ($\varepsilon$=0.4%~5.3%) | 10000~500 ($\varepsilon$=0.4%~1.7%) |

**e** Chaffee-Infante equation

| True equation form |
| --- |
| $u_t - u_{xx} + u - u^3 = 0$ |

| Discovered equation |
| --- |
| $1.03u_t - u_{xx} + 0.95u - 1.00u^3 = 0$ |

| Noise robustness | Data robustness |
| --- | --- |
| 0%~75% ($\varepsilon$=4.6%~7.4%) | 10000~1000 ($\varepsilon$=0.9%~4.6%) |

**f** KG equation

| True equation form |
| --- |
| $u_{tt} - 0.5u_{xx} + 5u = 0$ |

| Discovered equation |
| --- |
| $u_{tt} - 0.498u_{xx} + 5.04u = 0$ |

| Noise robustness | Data robustness |
| --- | --- |
| 0%~100% ($\varepsilon$=0.6%~26%) | 10000~500 ($\varepsilon$=0.6%~3.4%) |

**g** Eq.(6.2.12)

| True equation form |
| --- |
| $0.1u_{xt} + u_t + 0.1u_x = 0$ |

| Discovered equation |
| --- |
| $0.115u_{xt} + 1.03u_t + 0.1u_x = 0$ |

| Noise robustness | Data robustness |
| --- | --- |
| 0%~50% ($\varepsilon$=7.4%~19%) | 10000~1000 ($\varepsilon$=7.4%~19%) |

**h** Eq.(8.14.1d)

| True equation form |
| --- |
| $u_t + u_x/x + 0.25u_{xx} = 0$ |

| Discovered equation |
| --- |
| $0.998u_t + 0.981u_x/x + 0.25u_{xx} = 0$ |

| Noise robustness | Data robustness |
| --- | --- |
| 0% ($\varepsilon$=0.95%) | 10000~100 ($\varepsilon$=1.0%~1.9%) |

**Fig. 2. The performance of the proposed generative framework in discovering canonical PDEs with sparse and noisy data. (a)** Korteweg-De Vries (KdV) equation. **(b)** Burgers' equation. **(c)** Convection–diffusion equation. **(d)** Wave equation. **(e)** Chaffee–infante equation. **(f)** Klein–Gordon (KG) equation. **(g)** Eq. (6.2.12) in the math handbook[26]. **(h)** Eq. (8.14.d) in the math handbook. $\varepsilon$ is the coefficient error.

## Comparison against existing PDE discovery algorithms

To better elucidate the performance of our proposed framework, a comparative analysis was conducted against existing PDE discovery algorithms. Herein, we focus on comparing the discovery scope of PDEs and discovery efficiency. Acknowledging the diversity of existing PDE discovery methods, eight representative algorithms are selected for comparison: PDE-FIND[6], the generalized genetic algorithm (GGA)[34], DISCOVER[12], PDE-READ[29], W-SINDy[8], LLM4ED[17], and WeakIdent[9]. To highlight the inherent differences between algorithms, the PDE discovery process was conducted on the metadata generated by the trained surrogate model, thereby avoiding the influence of data processing. For a fair comparison, the candidate library in sparse regression-based methods includes the combination of polynomial terms and high-order derivatives, and all relevant vocabularies consistent with those employed in the proposed method. This results in a comprehensive candidate library comprising 43 candidate terms. For EqGPT, the correct PDEs are excluded from the PDE dataset during the training process, thereby avoiding data leakage. EqGPT must rely solely on its learned representations of general PDE structures and term interactions to generate and optimize toward the correct form. The details of experimental settings are provided in Materials and Methods. Eight canonical PDEs are examined, which involves typically-used baselines, like Burgers' equation and the KdV equation, as well as complex-form PDEs recorded in the math handbook[26]. The ability of existing methods and our proposed framework to discover these PDEs and corresponding time overhead are provided in Fig. 3.

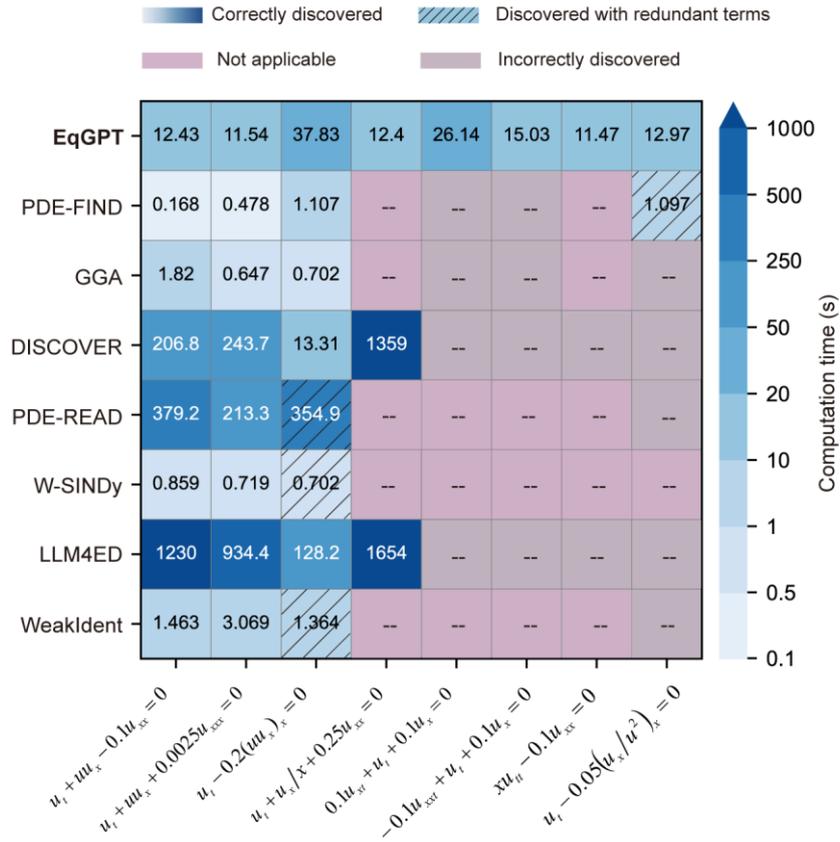

**Fig. 3. The comparison between the proposed method and existing methods on the discovery of eight canonical PDEs.** The values in the heatmap indicate the computational cost.

The result reveals notable differences in the discovery scope across various methods. Sparse regression methods, like PDE-FIND, W-SINDy, and WeakIdent, have the highest computational efficiency; however, the discovery scope is confined to identifying baseline equations with normal spatial and temporal derivatives. Even with a complete candidate library, these methods struggle to maintain sparsity when recovering PDEs with compound or nonstandard structures. Meanwhile, without special treatment[35], the terms, like $xu_{tt}$ and $u_x/x$, cannot be directly identified. In contrast, EqGPT can generate such terms by flexibly combining operators and variables, enabling the discovery of free-form PDEs beyond the limitations of predefined candidate libraries. This capability is essential for identifying new PDE structures in practice. Notably, the terms, like $u_{xt}$ and $u_{xxt}$, are difficult to be discovered when the left-hand-side (LHS) term is fixed as $u_t$, a limitation that EqGPT overcomes by not requiring the LHS term to be predetermined. While W-SINDy and WeakIdent employ weak-form formulations to enhance robustness and accuracy, these formulations are typically limited to polynomial terms and high-order derivatives, and are not well suited for capturing complex interaction terms.

Meanwhile, symbolic regression methods can discover free-form PDEs. However, the computational cost is extremely high, nearly 20 min for Eq. (8.14.1d). Its time overhead is even one thousand times larger than sparse regression methods. The low search efficiency also influences its ability to address more intricate terms. Although genetic algorithm-based methods, like GGA, can discover PDEs under simple compound form with higher computational efficiency, they are also incapable of handling complex terms that cannot be generated by cross-over and mutation. From the comparison, the LLM-based methods usually take a much larger computational time, which reflects the low efficiency of calling the general LLM by giving prompts. It is worth noting that while many symbolic regression models have demonstrated strong performance in prior studies [36,37], they typically lack support for differential operators and are therefore not directly comparable in PDE discovery tasks.

In the comparison against existing methods, our proposed framework successfully achieves a balance between the search scope and optimization efficiency. For baseline PDEs, the time overhead of EqGPT may be slightly higher than the sparse regression and genetic algorithms. However, the total time overhead is acceptable only with tens of seconds. Meanwhile, EqGPT presents stability and efficiency on the discovery of PDEs with complex temporal terms and intricate spatial terms. This capability is attributed to the integration of prior knowledge from established PDEs and the proposed GRE strategy, which enables the model to generate compact and plausible free-form PDEs, thereby facilitating more effective optimization during the discovery process.

**Identification of nonlinear dynamic systems in complex computed regions**
Boundary conditions and computed regions play important roles in simulating the behaviour of dynamic systems. In real-world scenarios, the computed regions are often complex and irregular. However, previous studies have not explored the discovery of PDEs within complex regions. Typically, earlier research focused on cases with regular regions, such as rectangular regions. Nevertheless, complex computed regions with irregular boundaries can amplify the challenge of discovering PDEs since the data complexity is escalated and the data continuity is reduced. In this analysis, we explore several scenarios featuring irregular computed regions to assess the performance of the proposed framework. Detailed information on these scenarios is provided in Supplementary Information S1.1. Poisson's equation with different boundary conditions is employed, which is written as:

$$\nabla^2 u = f \,.$$ (5)

If $f$=0, the equation is also called the Laplacian equation.

Initially, a simple disk region with radius $r$=1.5 is adopted. The Dirichlet boundary condition is $u(x,y)$=$x$sin($xy$). The solution dataset contains 40,200 observation data points on grids, as illustrated in Fig. 4a. The proposed framework is employed to discover the governing equation for the proof-of-concept. Notably, the target equation is deleted from the PDE dataset when training the generative model. Here, 30,000 discrete data points are randomly selected to train the surrogate model. The rewards of the top 10 equations in each optimization epoch are depicted in Fig. 4a. The reward of the best structure is dominantly higher than that of the other structures. The correct PDE structure has been discovered since the first epoch, which reflects the high optimization efficiency of the method. Furthermore, the precise alignment of the ultimately discovered PDE with the actual PDE attests to the method's remarkable accuracy.

Then, an irregular region, called the "smiley face region", is investigated. This region is carved out of the disk region with a radius of 2 to form a region of the smiley face. The Dirichlet boundary condition is as follows: $u(x,y)$=sin($x$+$y$). The solution dataset contains 44,711 observation data points on grids, as illustrated in Fig. 4b. Similarly, 30,000 discrete data points are randomly selected for PDE discovery. The rewards of the top 10 equations in each optimization epoch are depicted in Fig. 4b. The results of EqGPT are almost unaffected by this irregular region, exhibiting only a minor variation in the coefficients of the discovered PDE.

For a more difficult scenario, a computed region that consists of discontinuous glyph boundaries is considered. As illustrated in Fig. 4c, Poisson's equation with $f$=-1 is solved in the "EITech" region. For each alphabet, the zero Dirichlet boundary condition, where $u(x,y)$=0, is adopted. This complex scenario is simulated via *Mathematica*[38], and a dataset of 48,367 data points is obtained. A total of 45,000 discrete data points are randomly selected for PDE discovery, and the results are provided in Fig. 4c. The proposed framework can accurately identify the correct PDE structure and estimate the coefficients with overall precision. Although there is a relatively large deviation in the constant term $f$ of the PDE, the coefficient error remains below 2%. This finding indicates that the proposed framework can handle scenarios with complex discontinuous regions.

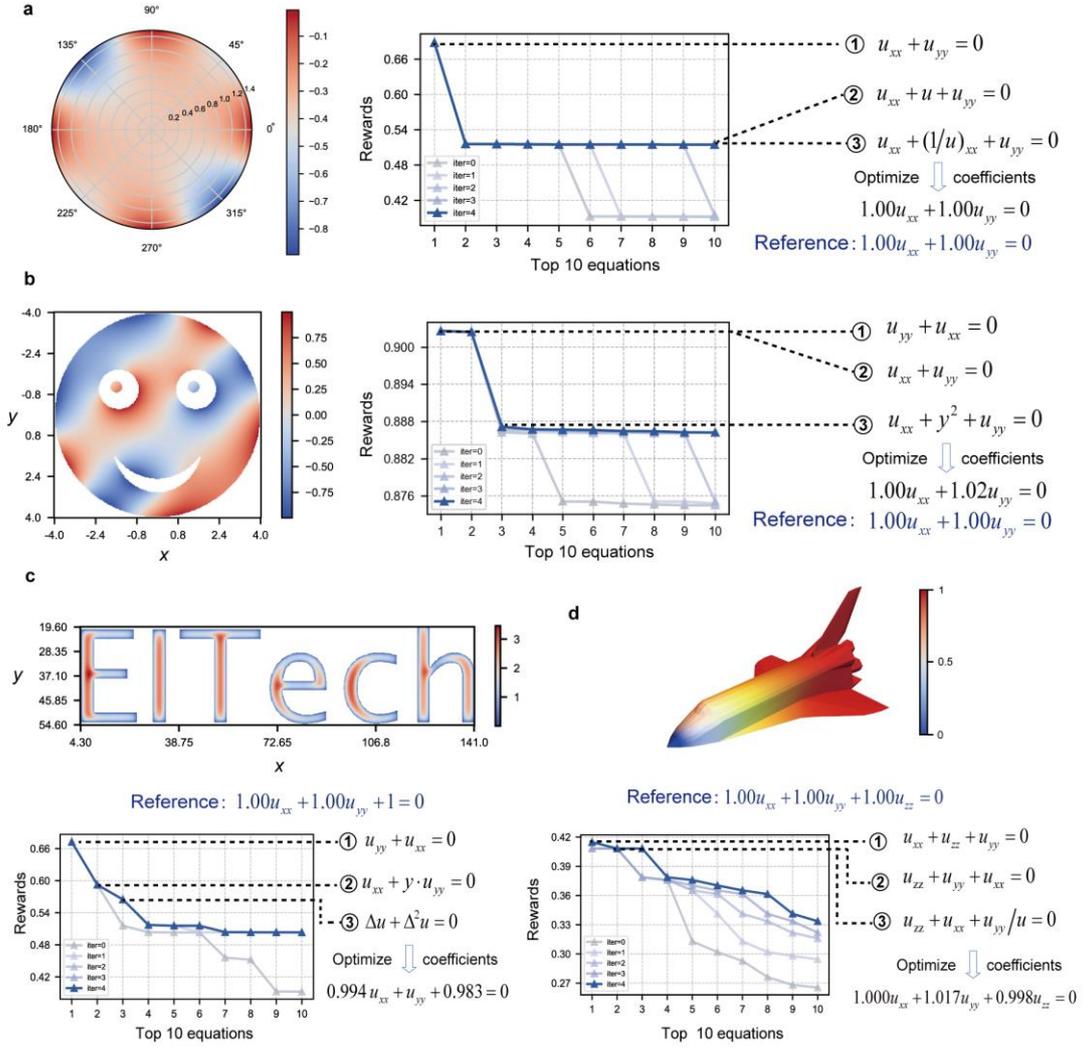

**Fig. 4. Discovery of Poisson's equation within complex computed regions. (a)** The dataset on a disk region (left) and the rewards of the top 10 equations generated from the EqGPT model in each optimization epoch (right). **(b)** The dataset on a "smiley face region" (left) and the rewards of the top 10 equations in each epoch (right). **(c)** The dataset on a discontinuous glyph boundary region of "EITech" (upper) and the rewards of the top 10 equations in each epoch (lower). **(d)** The dataset on the region of the 3-dimensional space shuttle (upper) and the rewards of the top 10 equations in each epoch (lower). A deeper blue colour indicates a larger epoch. The top 3 structures in the final epoch are listed.

Finally, a 3-dimensional (3D) scenario is investigated, in which the simulated region is a 3D space shuttle (Fig. 4d). The high-dimensional region is highly irregular and complex, which poses a great challenge to PDE discovery. Here, the simulation is conducted via *Mathematica*, and the Dirichlet boundary condition is as follows: $u(x,y,z)=1$ if $z \leq 1.3$ and $u(x,y,z)=0$ if $x \leq -7$. The dataset contains 438,960 data points, of which 40,0000 data points are employed for PDE discovery. The results are illustrated in Fig. 4d. The optimization process, while challenging, is ultimately effective. This is demonstrated by the gradual increase in the rewards of the top PDE structures with optimization epochs. The ultimately discovered PDE is closer to the true PDE, with a coefficient error of 0.63%. This substantiates the accuracy and stability of the proposed framework, even when confronted with high-dimensional complex regions.

## Extension to complex PDEs in high-dimensional space

The proposed framework is readily scalable to higher dimensions, given that the PDE dataset encompasses PDEs that range from one to three dimensions. Moreover, differential operators, such as the divergence operator ($\nabla$), Laplacian operator ($\Delta$), and even the BiLaplacian operator ($\Delta^2$), are considered and can be generated via the EqGPT model. These operators are applicable for all dimensions, which improves the flexibility of the proposed framework. By specifying the target dimension, EqGPT can generate proper PDEs in the respective dimension. Here, experiments are conducted to examine the extension of the EqGPT framework to high-dimensional scenarios. Details of the utilized dataset can be found in Supplementary Information S1.1.

First, the discovery of Burgers' equation in a 2-dimensional space is investigated, as depicted in Fig. 5a. This case serves as a typical benchmark for assessing the performance in uncovering high-dimensional PDEs. The true PDE form is written as follows:

$$u_t + (uu_x + uu_y) - 0.01(u_{xx} + u_{yy}) = 0 \qquad (6)$$

The dataset consists of the grid data of 101 spatial observation points in the domain $x \in [-1,1]$, 51 spatial observation points in the domain $y \in [-1,1]$, and 100 temporal observation points in the domain $t \in [0,2)$. The data size is 515,100. For PDE discovery, 200,000 data points are randomly selected to construct the surrogate model. The results of the proposed generative framework are illustrated in Fig. 5a. Evidently, it successfully discovers the correct PDE with a high accuracy of coefficients. Interestingly, the Laplacian operator ($\Delta u$) occurs in the discovered PDE, which diminishes the number of terms and leads to a more parsimonious PDE. The optimization cycle in the framework is also efficient, where the rewards of the top structures in the final epoch are much higher than those in the initial epoch. Additional experiments indicate that the surrogate model performs reliably in high-dimensional settings and remains robust under data noise levels of up to 15% (Table S2).

We then consider a more difficult high-dimensional scenario, which has never been discovered in prior works. The physical scenario is the vibration of an H-shaped elastic membrane, where the governing equation can be written as:

$$u_{tt} + u_{xx} + u_{yy} = 0 \, . \qquad (7)$$

The initial conditions are $u(0, x, y) = 2e^{-125(x-0.25)^2 + (y-0.5)^2}$ and $u_t(0, x, y) = 0$. The boundary condition is a zero Dirichlet boundary, $u(t,x,y)=0$. The dataset contains 3,774,981 data points, where 300,000 data points (7.9%) are randomly selected for PDE discovery. As illustrated in Fig. 5b, the physical field exhibits intricate temporal dynamics, posing significant challenges for PDE discovery. Despite this, the proposed framework remains successful in identifying the PDE structure, although there are some deviations in the coefficients. Considering that only 7.9% of the total data points are utilized, the results are acceptable.

Finally, we consider a practical case study involving a three-dimensional multiphase flow system with multiple state variables. The scenario simulates an oil–water two-phase displacement process, where water is injected at the two corners on the left boundary, and oil is produced from the center of the right boundary (Fig. 5c). This physical process is governed by a system of coupled PDEs that separately describe the dynamics of the oil and water phases, which are written as:

$$\phi \frac{\partial S_w}{\partial t} - \alpha \nabla [K_w \cdot \nabla (P_w - \rho_w \mathbf{g})] = 0 \tag{8}$$

$$\phi \frac{\partial S_o}{\partial t} - \alpha \nabla [K_o \cdot \nabla (P_o - \rho_o \mathbf{g})] = 0 \tag{9}$$

where $\phi$=0.1 denotes the porosity; the coefficient $\alpha$=0.0085; $S_w$ and $S_o$ are saturation fields for water and oil, respectively; $P_w$ and $P_o$ are pressure fields; $K_w$ and $K_o$ are the effective permeability fields that vary spatially and temporally; the gravitational acceleration $\mathbf{g}$ acts only in the vertical ($z$) direction and is set to 0.098; the water density $\rho_w$ is set to be 1; the oil density $\rho_o$ is 0.8; the two equations are coupled through the saturation constraint, $S_w+S_o$=1; and under the assumptions of incompressibility and negligible capillary pressure, the oil phase pressure equals the water phase pressure, i.e., $P_o=P_w$. Therefore, the PDE system can be simplified as:

$$\frac{\partial S_w}{\partial t} - 0.085 \nabla (K_w \cdot \nabla P_w) + 0.008 \frac{\partial K_w}{\partial z} = 0 \tag{10}$$

$$\frac{\partial S_w}{\partial t} + 0.085 \nabla (K_o \cdot \nabla P_w) - 0.007 \frac{\partial K_o}{\partial z} = 0 \tag{11}$$

This physical system involves two primary state variables ($S_w$ and $P_w$) and two dependent state variables, as the effective permeabilities of the water and oil phases are correlated with the corresponding saturation fields. This reflects the complex and high-dimensional characteristics of real-world subsurface flow systems. Datasets are generated from the unconventional oil and gas simulator (UNCONG)[39]. The spatial grid dimensions are $n_x$=50, $n_y$=50, and $n_z$=8, with the simulation time ranging from day 0 to day 30, discretized into 30 time steps. Each physical variable yields 600,000 data points. During PDE discovery, potential terms associated with $S_w$, $K_w$, and $K_o$ are incorporated into the vocabulary, while terms related to $P$ are constructed analogously to the single-variable setting. Spatial symmetry in the $x$ and $y$ directions is utilized to constrain the search space for the generated equations. All available data are used for the discovery of governing equations. Additional details about the experimental setting are provided in Supplementary Information S1.1. The PDE discovery process involves 600 populations with 5 optimization epochs. The resulting discovered PDEs are written as:

$$\frac{\partial S_w}{\partial t} - (0.095 \nabla K_w \cdot \nabla P_w + 0.121 K_w \cdot \Delta P_w) + 0.009 \frac{\partial K_w}{\partial z} = 0 \tag{12}$$

$$\frac{\partial S_w}{\partial t} + (0.089 \nabla K_o \cdot \nabla P_w + 0.085 K_o \cdot \Delta P_w) - 0.009 \frac{\partial K_o}{\partial z} = 0 \tag{13}$$

Notably, the identified equations correspond to the expanded form of the ground-truth PDEs (Fig. 5c). The true governing equations include nested differential operators, which are difficult to recover directly without explicit prior knowledge. However, the proposed method effectively reconstructs these structures by generating free-form interaction terms. This result suggests that the framework is capable of unveiling underlying relationships among multiple state variables, even in three-dimensional space with time. Although the numerical coefficients of the discovered equation show slight deviations from the ground truth, the discrepancies remain within acceptable bounds given the complexity of the physical processes involved.

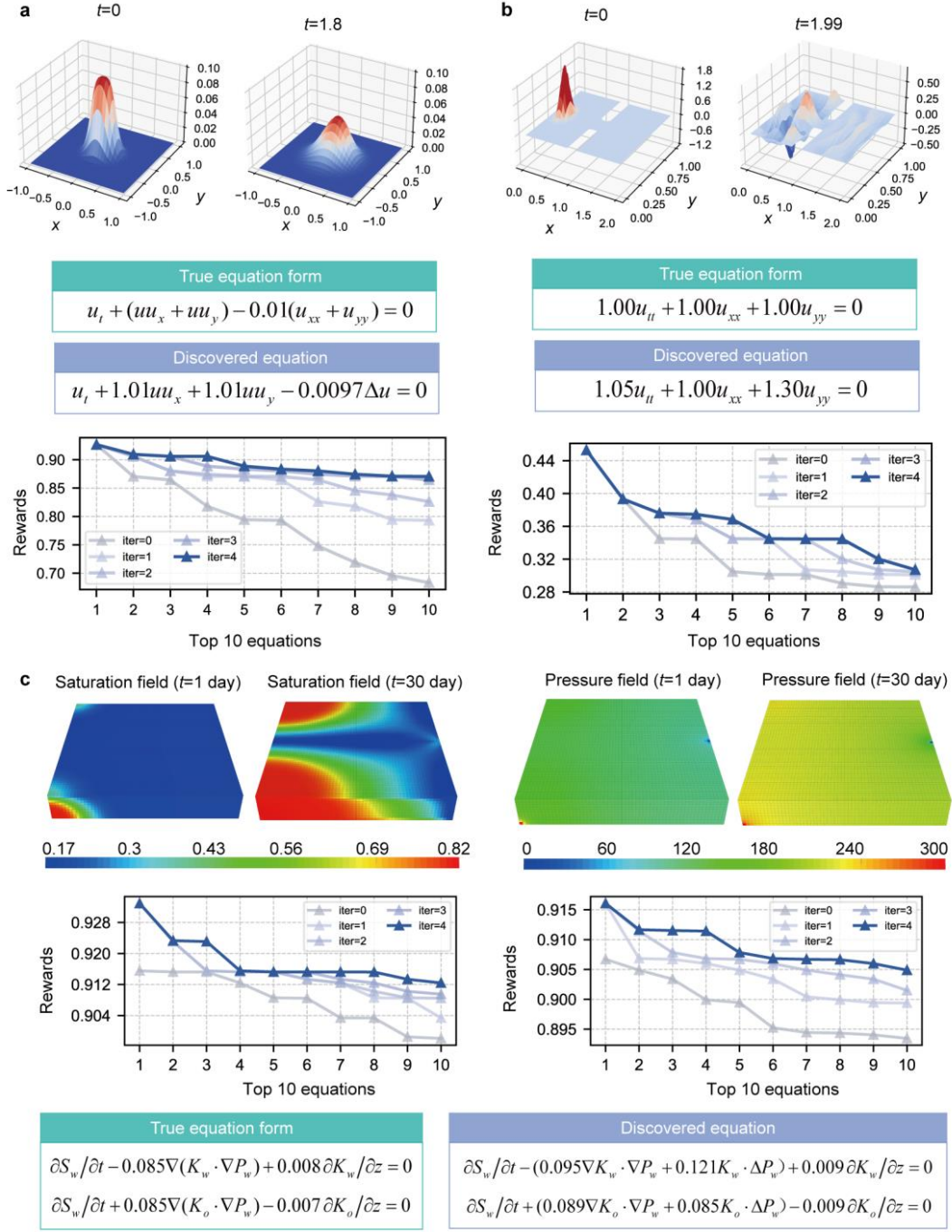

**Fig. 5. Discovery of high-dimensional PDEs via the proposed generative framework. (a)** For the 2-dimensional (2D) Burgers' equation, the surface of $u(x,y)$ when $t=0$ and $t=1.8$, the true PDE, the discovered PDE, and the rewards of the top 10

equations in each epoch. **(b)** For the vibration of an H-shaped elastic membrane, the physical field when $t$=0 and $t$=1.99, the true PDE, the discovered PDE, and the rewards of the top 10 equations in each epoch. **(c)** The saturation field and the pressure field of water when $t$=1 day and 30 day in the 3-dimensional oil–water two-phase displacement process. The true PDE, the discovered PDE, and the rewards of the top 10 equations when discovering the governing equation of water (left) and oil (right) phases in each epoch.

### Discovery of new PDEs from real-world experimental data

PDE discovery has long been anticipated to uncover previously undisclosed governing equations from actual experimental data. Therefore, we aim to evaluate the proposed method's ability to discover new equations and principles from real-world experimental data. As a practical example, we consider the discovery of the governing equation for highly nonlinear surface gravity waves propagating toward breaking. Despite the ubiquity of wave breaking in natural environments, its detailed physical and statistical mechanisms remain incompletely understood, making the modelling of such processes a significant challenge. To this end, we focus here on describing the evolution of unidirectional, irregular focused wave groups as they approach the onset of breaking. Particularly, there is currently no theoretical governing equation based on surface elevation at the air-water interface to describe this process towards an irreversible process, which hinders a deeper understanding and necessitates the data-driven discovery of potential governing equations.

The real-world data employed in this study are subsets from two experimental campaigns, namely BUBER and EURUS. Briefly, both campaigns were carried out in a 27.2-meter-long, glass-walled wave tank located at the Department of Civil and Environmental Engineering, Imperial College London. Breaking wave groups were generated using a bottom hinged wave paddle at one end of the tank, following the so-called dispersive-focusing technique and characterised by JONSWAP-type wave spectra.[40,41] Three high-resolution, charge-coupled device (CCD) cameras were placed outside of the flume to capture the propagation process over a spatial domain spanning approximately 8 meter to 12.5 meter (Fig. 6a), with a frame rate of 20 Hz. The spatial surface elevation was reconstructed at different time instants before the onset of breaking by employing the image-processing technique developed by Cao et al.[42] on a frame-by-frame basis, allowing for the air-water boundary to be identified (Fig. 6b). Due to visual occlusion caused by structural columns of the wave tank, the fields of view of the three cameras were spatially separated and did not overlap. As a result, Fig. 6b contains three distinct, non-contiguous spatial regions. Nevertheless, no explicit interpolation or reconstruction was applied to address the spatial gaps, as this was deemed beyond the scope of the present research. A total of 12 independent experiments were conducted with different initial conditions of the wave paddle, including the peak enhancement factor and peak wave period of the JONSWAP-type spectra, as well as the total amplitude summed across all underlying wave components. The total datapoints are 4,583,721 with approximately 380,000 for each experiment. Additional details about the experimental setting are provided in Supplementary Information S1.1.

In the PDE discovery process, the data were non-dimensionalized:

$$x^* = (x - x_0)/\lambda; \ t^* = t/T_p; \ \eta^* = \eta/\lambda \qquad (14)$$

where $x_0$ is the beginning coordinate of the observed region; $\lambda$ is the initial wavelength; $T_p$ is the initial peak period; and $\eta$ is the surface elevation. The surrogate models are constructed for data from each experiment, and standard EqGPT is applied to discover the underlying governing equation. For generated PDEs, the mean of reward calculated on the data from each independent experiment is utilized for evaluation. The optimization process is shown in Fig. 6c. The final discovered equation is:

$$\eta_t^* = c_1 \eta_x^* + c_2 \eta_{xxx}^* + c_3 (\eta^{*2})_{xxx} . \qquad (15)$$

For different initial conditions, the coefficients may be slightly different (Table S3). As shown in Fig. 6c, this PDE performs well across all experiments. It is discovered that the left-hand side term ($\eta_t^*$) and the regressed right-hand side terms of the PDE align well at the time of wave breaking onset (Fig. 6d). The time-marching posterior prediction by the discovered PDE also closely matches the observed data, demonstrating its accuracy in representing this physical process. Notably, our method successfully uncovers governing equations under partial information since real-world experimental data are incomplete and contain missing regions.

From the analysis of fluid mechanics, the discovered PDE is physically reasonable, which incorporates the $\eta_x^*$ term as the leading order propagation term. This is consistent with the formal mathematical analysis, which shows that the same leading order term with $\eta_x^*$ can be derived for non-breaking water waves under the potential flow framework[43]. This indicates that the evolution of bulk water towards the breaking onset still mostly follows the classic wave theory, which agrees well with the observation that wave breaking is a strong but localised behavior. Apart from the classic term, we further observed a new functional form as $(\eta^{*2})_{xxx}$, where the triple partial $x$ derivative can be found in the classic shallow water equation – KdV equation. The new form, however, instead of being a linear dispersive term, can give rise to harmonics at higher frequencies as the wave evolves towards a breaking event. This is superficially similar to the wave-breaking behaviors observed in previous studies, where the breaking events are found to be associated with higher energies at higher frequencies. [44,45]

Interestingly, this term occurs in the PDE dataset collected from the math handbook within a chapter introducing compactons, which was originally concerned with mathematical properties rather than physical applications. This finding reflects the effectiveness of our proposed strategy: the broad learning of PDE forms across diversified domains is beneficial for discovering new equations in specific fields. Absent such cross-domain knowledge, such a higher-order nonlinear term would be difficult to conceive and include in traditional methods' candidate libraries. This example demonstrates the importance of knowledge-guidance, and confirms the potential of our method to discover new principles and advance scientific understanding in real-world problems.

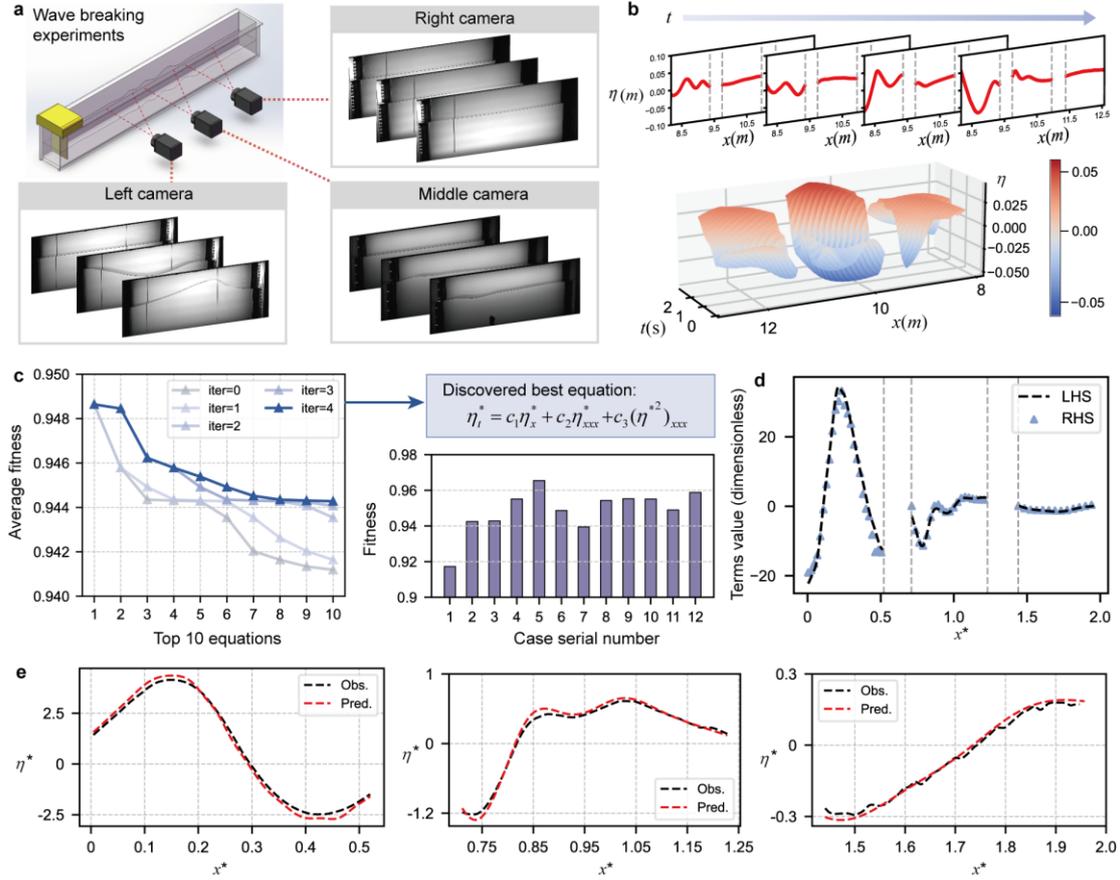

**Fig. 6. Discovery of previously undisclosed governing equations for highly nonlinear surface gravity waves propagating toward breaking from real-world experimental data. (a)** Experimental setup and the original records from the high-resolution, charge-coupled device cameras. The diagram is not to scale. **(b)** The visualization of the recognized water-air boundary. The blank areas between the grey dashed lines represent regions that could not be captured due to obstruction by columns in the tank. **(c)** The discovered equation, the visualization of the optimization process, and the fitness in 12 independent experiments. **(d)** The comparison between the left-hand-side (LHS) term and the regressed right-hand-side term (RHS) of the discovered equation. **(e)** The observation and posterior prediction from the discovered governing equation in different regions.

## Discussion

In this study, we explored a knowledge-guided approach for the discovery of partial differential equations (PDEs), representing a step toward integrating data-driven methodologies with domain-informed generation. This framework addresses a longstanding dilemma of balancing both search space and optimization efficiency in purely data-driven approaches. By incorporating prior knowledge extracted from mathematical handbooks, the proposed method autonomously generates and optimizes free-form PDEs with improved accuracy and robustness, even in challenging scenarios involving sparse or noisy data, complex computed regions, and high-dimensional domains. The practical example of discovering a new governing equation from real-world experiments demonstrates the practical utility of the method and underscores the core principle of this work: "stones from other hills may polish one's own jade", i.e., extensive learning from existing equations can facilitate the identification of new governing laws. Further ablation studies confirm the positive impact of increasing the number and diversity of embedded PDE forms on model performance (Supplementary Information S1.3).

Previous studies have attempted to construct generative models for symbolic regression[13,22,46]; however, these models largely focused on algebraic expressions and lacked the capacity to handle differential operators, making them unsuitable for PDE discovery. Unlike general symbolic regression, the complexity of PDEs arises not from tree depth or structural complexity, but from the combinatorial interactions of differential terms. While tree-based representations offer expressive power, they also introduce significant computational overhead and pose optimization challenges. In this work, we observe that despite the structural diversity of PDEs, their constituent elements exhibit regularities that can be quantified and encoded. To that end, we introduce a generative representation of equations (GRE) strategy, in which PDE structures are reformulated as sentence-like sequences of operators and basic terms. This compact representation reduces the search space while preserving expressiveness, enabling efficient generation of free-form PDEs. By embedding known equation forms into the learning process, the model not only identifies plausible terms, but also learns their co-occurrence patterns, thereby guiding the discovery process in a more informed and structured manner.

Nonetheless, this study has certain limitations. The model's performance is influenced by the quality and coverage of the training data derived from existing PDE sources. Expanding the dataset to include equations from a wider range of scientific domains may further enhance generalizability and reduce domain-specific biases. Future work will focus on enriching the PDE dataset and improving the model's adaptability to broader classes of physical systems. Given the stability, precision, and scalability of the proposed generative framework, it is promising for practical application in PDE discovery within the scientific community.

## Materials and Methods

**The construction of the PDE dataset from the math handbook.** In this work, a dataset of PDE forms is established from the scientific handbook *Nonlinear Partial Differential Equations*[26]. In this handbook, a variety of common PDEs, which are commonly utilized in science and engineering, are introduced and analysed in detail. Here, PDEs that contain explicit terms are recorded in the PDE dataset. Each PDE in the dataset is recorded solely based on its structure, which means that PDEs with identical structures, but different coefficients, are considered identical within the dataset. Moreover, the equivalent forms of PDEs are differentiated by experts, which avoids repetition and improves the representativity of the dataset. PDEs are recorded in the form of a structure string, which is separated by commas. The conversion from mathematic form to structure string is currently accomplished manually, since humans can easily understand and separate the basic components in the PDE. In this work, 221

PDEs of different structures are recorded in the dataset. These PDEs can be decomposed and summarized into 56 basic components (i.e., vocabularies), including terms and operators, which serve as the foundation for the subsequent implementation of scientific-augmented training (SAT).

Considering that the sequence of terms in PDEs does not affect their essence, a data augmentation technique is adopted in this work to expand the size of the dataset for training models. In data augmentation, the commutative properties of addition and multiplication are leveraged to generate equivalent PDE structures simultaneously. The first method involves swapping the sequence within the terms. For example, the term $u \times u_x$ equals $u_x \times u$, while the encoded vectors are different. The other is swapping the sequence between terms. For example, the structure $u \times u_x + u_{xx}$ equals $u_{xx} + u \times u_x$. By combining the two methods, all equivalent forms of the PDE can be obtained. Programming can be accomplished by identifying the operator of $\times$ and $+$ for exchanging the corresponding sequence. The PDE dataset consists of 221 different PDEs collected from math handbooks. To enable generative representation of equations, each PDE is expressed as a sentence-like structure composed of operators and terms. Since the semantic meaning of a PDE remains invariant under permutation of additive terms, for example, $u_t + u_x + u_{xxx} = 0$ is equivalent with $u_{xx} + u_t + u_x = 0$, we leverage this property to perform data augmentation. Specifically, for each PDE form, we generate 32 distinct sentence variants by randomly shuffling the order of additive terms. In cases in which a PDE admits more than 32 equivalent permutations, a random subset of 32 is selected; for PDEs with fewer permutations, additional samples are generated through random augmentation to reach a total of 32 sentence representations per equation. As a result, we obtain a training dataset comprising 7,072 sentence-style entries that encode mathematical prior knowledge, which is subsequently used to train the EqGPT.

**Generative representation of equations (GREs).** The proposed GRE consists of a pivotal component within the framework, where a specialized method for representing PDEs is adopted. In the featurization procedure, vocabularies are first represented via sequential encoding. In addition, the start token (S) and end token (E) are also encoded to indicate the starting and ending of the sentence, respectively. Specifically, E is encoded to be 1, + is 2, $\times$ is 3, / is 4, S is 5, and terms of the PDE are encoded sequentially after then. This encoding method separates the operators and terms, which facilitates the subsequent training and generating process. In this way, the PDEs can be converted into vectors. When training the EqGPT model, the PDEs are further transformed into a matrix through one-hot encoding of each vocabulary. The GRE establishes a direct correspondence between equations and vectors, enabling efficient interpretation of the generated vectors. As a result, the output vectors can be quickly decoded back into their corresponding PDE forms.

**The design of the EqGPT generative model.** In the proposed PDE discovery framework, a generative model, called EqGPT, is established to learn from the PDE dataset and generate new PDEs. Its architecture is similar to that of GPT-2, which consists of multiple transformer blocks. The input training data are embedded into high-dimensional spaces through the embedding layer, including both token embeddings and positional embeddings. Then, the inherent features and co-occurrence probabilities between vocabularies in the PDE structure are extracted and learned in blocks of the transformer. The transformer is composed of a layer norm and a masked multi-head self-attention layer. The details of the implementation can be found in the literature[27]. The output of the model is the probability distribution of the next word, which has been processed by the Softmax function.

To support the design of EqGPT, we draw upon the theoretical foundations of autoregressive language models, particularly those instantiated by transformer-based architectures, such as GPT-2. At the core of these models lies the principle of next-token prediction conditioned on preceding context, which enables them to capture complex sequential dependencies in tokenized inputs. In the context of PDE discovery, we treat symbolic expressions of differential equations as sequences composed of discrete semantic units. This formulation allows us to repurpose the autoregressive modeling framework for learning the co-occurrence patterns of PDE terms. Specifically, EqGPT learns to model the conditional probability distribution of the next token $x_i$ given the preceding tokens $x_1,...,x_{t-1}$, i.e., $P(x_i|x_1,...,x_{t-1})$. This co-occurrence learning is facilitated by the self-attention mechanism, which dynamically weights context tokens according to their relevance to the prediction target. Although the EqGPT model is trained on a relatively small dataset compared to its parameter count, this overparameterization does not necessarily lead to overfitting.[47] On the contrary, large models can exhibit better generalization by adapting their internal representations more flexibly, especially when paired with techniques, such as dropout, weight decay, and appropriate regularization. Moreover, the autoregressive nature of EqGPT enables it to generate new equation structures in a free-form yet syntactically valid manner, by sampling from the learned token distributions.

**Scientifically augmented training (SAT)**. The SAT enables the EqGPT model to extract and utilize scientific knowledge from mathematical handbooks, thereby increasing its generative capacity and empowering it to produce PDE structures that adhere to established scientific principles. The sentence of the PDE is separated into different lengths of preceding fragments (tokens) for training. The training loss is the cross-entropy loss of the predicted next word from preceding tokens and the true word. The optimizer is Adam[48], and the learning rate is $10^{-4}$. The batch size is 128, and the number of training epochs is 100. In the generation process, the start token (S) is first input into the trained model, and the next token is sampled from the predicted probability distribution. Here, a combination of probabilistic sampling and random sampling is adopted to improve the diversity of the generation. Specifically, several scientific constraints are imposed to guarantee the validity of the generated PDEs in the SAT process. First, the terms and operators are considered to occur at intervals. Specifically, if the preceding token is an operator (index<6), the next token must be a PDE term (index $\geq$ 6). Therefore, the probability distribution can be constrained according to the predicted location. For example, when the length of the input is odd, only the probabilities of terms are reserved, whereas the others are forced to be zero. Similarly, when the length of the input is even, only the probabilities of the operators are reserved. In this way, the PDE form is guaranteed to be scientifically proper, which avoids the occurrence of consecutive operators or terms. Furthermore, the predicted probability is tailored according to the input variables or dimensions. For example, if the input variables are $x$ and $t$, the probabilities regarding $y$ and $z$ are forced to be zero. By employing this scientifically augmented method, the probability distribution can be automatically calibrated to accommodate problems of different dimensions.

When selecting the next token from the constrained probability distribution, the combination of probabilistic sampling and random sampling is adopted in the SAT. In addition, under a certain probability (20% in this work), the token is chosen by random selection of all possible valid tokens. This treatment enables the model to explore more diverse possibilities. When the end token (E) is sampled or the length of the output reaches a threshold (50 in this work), the generation process stops. The sentence between the start token and the end token is the generated PDE structure. Notably, in the generation process, even for the same input, the outputs can be different.

**The optimization cycle for PDE discovery**. The generative model, after being trained through SAT, is capable of generating a diverse range of valid PDE structures, thereby providing a broad yet effective search space. An optimization cycle is constructed to fine-tune the generative model to realize the directed generation of PDEs that conform to the observation data. As illustrated in Fig. 1, the observation data, which may be sparse and noisy, are utilized to train a fully connected artificial neural network (ANN) that functions as a surrogate model. The ANN consists of 7 layers, including an input layer, 5 hidden layers, and an output layer. The input neurons are equal to the number of dimensions of the problem, the output neuron is 1, and the number of neurons in the hidden layer is 50. The activation functions are Sin or Rational functions[29]. The training epoch is 50,000, and the early stopping technique is adopted to avoid overfitting. In practice, for sparse and noisy datasets, training typically converged and terminated within 1,000 to 2,000 epochs, well before reaching the predefined limit. During this process, the validation error is monitored, and the epoch exhibiting the lowest validation error is identified as the optimal one.

After the surrogate model has been trained, abundant meta-data are generated on grids from model predictions, where the data noise is smoothed, and derivatives are calculated through automated differentiation. These meta-data are utilized to calculate the reward of the generated PDEs from the generative model. In this work, EqGPT generates 400 PDEs in each optimization cycle. The reward is calculated via Eq. (1). Notably, EqGPT only generates the structure of PDEs, and the coefficients are obtained from the least squares regression via meta-data. The calculated rewards are then sorted, and the top 10 PDEs with the highest rewards are selected as the inputs for fine-tuning the generative model (Fig. S1). In the fine-tuning process, the learning rate is $10^{-5}$, and the training epoch is 5. This means that fine-tuning will only slightly optimize EqGPT, increasing its propensity to generate PDE structures that closely resemble these top PDEs. Afterwards, the fine-tuned EqGPT model continues to generate 400 new PDEs, and the top 10 PDEs are selected similarly. Out of the top 10 equations from both the new and old rounds, totalling 20 equations, 10 equations with the highest rewards are selected to advance to the subsequent optimization round. The optimization cycle stops when the maximum number of iterations is reached, which is set to 5 in this work. The best structure in the last iteration is the discovered optimal PDE. Notably, the current framework can also provide alternative suboptimal structures by sorting the rewards in the final iteration. In terms of the computational cost, the generation process is fast, i.e., only seconds, and each optimization cycle takes 10~30 s on the CPU i9-14900 KF and GPU RTX 4090D.

When the optimal PDE structure has been discovered, the coefficients can be calculated through least squares regression and further optimized by a physics-informed neural network (PINN)[49]. In the optimization of coefficients, the discovered PDE is incorporated into the previously trained surrogate model to construct a PINN. In each training epoch of the PINN, the derivatives are calculated through automatic differentiation, and the coefficients of the PDE are updated via least squares regression. The loss of the PINN can be calculated as:

$$
\begin{aligned}
L_{PINN} &= \lambda_{Data} MSE_{Data} + \lambda_{PDE} MSE_{PDE} \\
&= \frac{\lambda_{Data}}{N} \sum_{i=1}^{N} (u - u_{pred})^2 + \frac{\lambda_{PDE}}{N} \sum_{i=1}^{N} \left( \theta_{LHS} - \vec{\xi} \theta_{RHS} \right)^2
\end{aligned}
\tag{16}
$$

where $N$ is the number of training data; $MSE_{Data}$ is the data loss, which is the same as that of the ANN; and $MSE_{PDE}$ is the mean squared error calculated from the least squared regression of the PDE. In this work, $\lambda_{Data}$ is set to 1, and $\lambda_{PDE}$ is set to 0.01. Then, the PINN is optimized by the loss function. The maximum optimization epoch for the PINN is 500, and the calculated coefficients in the final epoch are the optimal coefficients. Through PINN optimization, the accuracy of the discovered PDEs can be further improved[34].

**The experimental settings.** In this work, some experiments are conducted for proof-of-concept. In the discovery of canonical PDEs with sparse and noisy data, the number of meta-data is 10,000, and the input variables are $x$ and $t$. When discovering each PDE, the target structure is deleted from the dataset, and the EqGPT model is retrained by the modified dataset. This guarantees that the true structure has never been trained by the generative model. The penalty strength $\alpha_0$ in Eq. (1) is 0.2.

In the comparison with existing PDE discovery methods, all of the experiments are conducted on the meta-data generated from the surrogate model. For the sparse-regression based algorithm, the candidate library comprises the combination of 3rd-order polynomial terms and up to 4th-order derivatives, and all relevant vocabularies consistent with those employed in the proposed method. Considering that the target canonical PDEs are both in one-dimensional (1D) space, the vocabularies relevant to the variable $y$ and $z$ are omitted. Therefore, this results in a comprehensive library comprising 43 candidate terms.

$$
\Phi = [u, u_x, u_{xx}, u_{xxx}, u_{xxxx}, \ldots, u^3 u_{xxxx}, u_{xt}, u_{xxt}, x, t, \ldots, (uu_x)_x, (u_x/u^2)_x]
\tag{17}
$$

The maximum number of iterations is 250, $\lambda$ is 2, and $d_{tol}$ is 0.1. The $l_0$ penalty strength ranges from 0 to 1 in increments of 0.05. For the GGA algorithm, the basic genes are defined as $u$, $u_x$, $u_{xx}$, and $u_{xxx}$, the number of populations is 400, the maximum generation of evolution is 200, the mutation rate is 0.3, and the $l_0$ penalty strength is $3 \times 10^{-4}$. For the DISCOVER algorithm, the considered operators are plus, minus multiply, and divide. The threshold of the reserved expressions is 0.02, the coefficient of entropy loss is 0.03, the coefficient of policy gradient loss is 1, the maximum number of function terms is 6, the total number of generated expressions at each iteration is 500, and the maximum number of iterations is 100. For other algorithms, the parameter setting is the same as the codes provided in the open source.

For the discovery of PDEs in complex regions, the input variables are $x$ and $y$ for 2D cases and $x$, $y$, and $z$ for 3D cases. The penalty strength $\alpha_0$ is 1. Given the irregularity of the regions in these instances, the meta-data are generated at each observation point, excluding the boundaries. For the discovery of PDEs with high dimensions, the input variables are $x$, $y$, and $t$. The penalty strength $\alpha_0$ is 1.

**Acknowledgements**

This work was supported and partially funded by the National Natural Science Foundation of China (grant no. 52288101), the China Postdoctoral Science Foundation (grant no. 2024M761535), and the Natural Science Foundation of Ningbo, China (grant no. 2023J027). The experimental results used here were funded by the UK Natural Environment Research Council (grant no. NE/T000309/1) awarded to A.H.C. This work was supported by the High Performance Computing Centers at the Eastern Institute of Technology, Ningbo, and the Ningbo Institute of Digital Twin. The authors express gratitude to Prof. Michael Rogers for his selfless assistance in simulating PDEs via Mathematica.

**Data availability**

The dataset generated in this study has been deposited in the GitHub repository, https://github.com/woshixuhao/EqGPT/tree/main/data. Source data are provided with this paper.

**Code availability**

All of the original code has been deposited at the website, https://github.com/woshixuhao/EqGPT/tree/main/code.

**Author contributions**

H.X., Y.C., and D.Z. conceived the idea, designed the study, and analyzed the results. H.X. developed the algorithm, performed the computations, and generated the results and figures. M.D. participated in the investigation. A.C. obtained funding for and conceived the real-world experiments at Imperial College London. R.C., supervised by A.C., conducted the experiments. R.C., A.C., and T.T. analyzed the discovered equations from fluid mechanisms. J.L. performed the two-phase-flow simulations. Y.C. and D.Z. supervised the project. All of the authors contributed to the writing and editing of the manuscript.

**Declaration of interests**

The authors declare that they have no competing interests.

# References

1.  Brunton, S. L. & Kutz, J. N. Methods for data-driven multiscale model discovery for materials. *Journal of Physics: Materials* **2**, 044002 (2019).

2.  Zanna, L. & Bolton, T. Data-driven equation discovery of ocean mesoscale closures. *Geophys Res Lett* **47**, (2020).

3.  Beetham, S., Fox, R. O. & Capecelatro, J. Sparse identification of multiphase turbulence closures for coupled fluid-particle flows. *J Fluid Mech* **914**, (2021).

4.  Brunton, S. L. & Kutz, J. N. Promising directions of machine learning for partial differential equations. *Nat Comput Sci* **4**, 483–494 (2024).

5.  Brunton, S. L., Proctor, J. L., Kutz, J. N. & Bialek, W. Discovering governing equations from data by sparse identification of nonlinear dynamical systems. *Proc Natl Acad Sci U S A* **113**, 3932–3937 (2016).

6.  Rudy, S. H., Brunton, S. L., Proctor, J. L. & Kutz, J. N. Data-driven discovery of partial differential equations. *Sci Adv* **3**, 1–7 (2017).

7.  Stephany, R. & Earls, C. Weak-PDE-LEARN: A weak form based approach to discovering PDEs from noisy, limited data. *J Comput Phys* **506**, 112950 (2024).

8.  Messenger, D. A. & Bortz, D. M. Weak SINDy for partial differential equations. *J Comput Phys* **443**, 110525 (2021).

9.  Tang, M., Liao, W., Kuske, R. & Kang, S. H. WeakIdent: Weak formulation for identifying differential equation using narrow-fit and trimming. *J Comput Phys* **483**, 112069 (2023).

10. Xu, H., Chang, H. & Zhang, D. DLGA-PDE: Discovery of PDEs with incomplete candidate library via combination of deep learning and genetic algorithm. *J Comput Phys* **418**, 109584 (2020).

11. Chen, Y., Luo, Y., Liu, Q., Xu, H. & Zhang, D. Symbolic genetic algorithm for discovering open-form partial differential equations (SGA-PDE). *Phys Rev Res* **4**, (2022).

12. Du, M., Chen, Y. & Zhang, D. DISCOVER: Deep identification of symbolically concise open-form partial differential equations via enhanced reinforcement learning. *Phys Rev Res* **6**, (2024).

13. Sun, F., Liu, Y., Wang, J.-X. & Sun, H. Symbolic Physics Learner: discovering governing equations via Monte Carlo tree search. *arXiv preprint arXiv:2205.13134* (2022).

14. Kamienny, P.-A., Lample, G., Lamprier, S. & Virgolin, M. Deep generative symbolic regression with monte-carlo-tree-search. in *International Conference on Machine Learning* 15655–15668 (PMLR, 2023).

15. Makke, N. & Chawla, S. Interpretable scientific discovery with symbolic regression: a review. *Artif Intell Rev* **57**, (2024).

16. Vaswani, A. Attention is all you need. *Adv Neural Inf Process Syst* (2017).

17. Du, M., Chen, Y., Wang, Z., Nie, L. & Zhang, D. Large language models for automatic equation discovery of nonlinear dynamics. *Physics of Fluids*, **36**(9), (2024).

18. Shojaee, P., Meidani, K., Gupta, S., Farimani, A. B. & Reddy, C. K. Llm-sr: Scientific equation discovery via programming with large language models. *arXiv preprint arXiv:2404.18400* (2024).

19. Li, Y. *et al.* Mllm-sr: Conversational symbolic regression base multi-modal large language models. *arXiv preprint arXiv:2406.05410* (2024).

20. Ye, Z. *et al.* Pdeformer: Towards a foundation model for one-dimensional partial differential

equations. *arXiv preprint arXiv:2402.12652* (2024).

21.   Sun, J., Liu, Y., Zhang, Z. & Schaeffer, H. Towards a foundation model for partial differential equations: multi-operator learning and extrapolation. *Physical Review E*, *111*(3), 035304, (2024).

22.   Chen, J. *et al.* Symbol: Generating flexible black-box optimizers through symbolic equation learning. *arXiv preprint arXiv:2402.02355*. (2024).

23.   Mundhenk, T. N. et al. Symbolic regression via neural-guided genetic programming population seeding. In *Proceedings of the 35th International Conference on Neural Information Processing Systems*, 24912–24923 (2021).

24.   Landajuela, M. et al. A unified framework for deep symbolic regression. In *Proceedings of the 36th International Conference on Neural Information Processing Systems*, 33985–33998 (2022).

25.   Valipour, M., You, B., Panju, M. & Ghodsi, A. SymbolicGPT: a generative transformer model for symbolic regression. *arXiv preprint arXiv:2106.14131*. (2021).

26.   Debnath, L. *Nonlinear Partial Differential Equations for Scientists and Engineers*. (Springer, 2005).

27.   Radford, A. *et al.* Language models are unsupervised multitask learners. *OpenAI blog* **1**, 9 (2019).

28.   Xu, H., Chang, H. & Zhang, D. Dl-pde: Deep-learning based data-driven discovery of partial differential equations from discrete and noisy data. *Commun Comput Phys* **29**, 698–728 (2021).

29.   Stephany, R. & Earls, C. PDE-READ: Human-readable partial differential equation discovery using deep learning. *Neural Networks* **154**, 360–382 (2022).

30.   Zhang, S. & Lin, G. Robust data-driven discovery of governing physical laws with error bars. *Proceedings of the Royal Society A: Mathematical, Physical and Engineering Sciences* **474**, (2018).

31.   Kaheman, K., Kutz, J. N. & Brunton, S. L. SINDy-PI: A robust algorithm for parallel implicit sparse identification of nonlinear dynamics: SINDy-PI. *Proceedings of the Royal Society A: Mathematical, Physical and Engineering Sciences* **476**, (2020).

32.   Fasel, U., Kutz, J. N., Brunton, B. W. & Brunton, S. L. Ensemble-SINDy: Robust sparse model discovery in the low-data, high-noise limit, with active learning and control. *Proceedings of the Royal Society A: Mathematical, Physical and Engineering Sciences* **478**, (2022).

33.   Xu, H., Zeng, J. & Zhang, D. Discovery of partial differential equations from highly noisy and sparse data with physics-informed information criterion. *Research* **6**, 1–13 (2023).

34.   Xu, H. & Zhang, D. Robust discovery of partial differential equations in complex situations. *Phys Rev Res* **3**, (2021).

35.   Rudy, S., Alla, A., Brunton, S. L. & Kutz, J. N. Data-driven identification of parametric partial differential equations. *SIAM J Appl Dyn Syst* **18**, 643–660 (2019).

36.   Petersen, B. K. *et al.* Deep symbolic regression: Recovering mathematical expressions from data via risk-seeking policy gradients. *arXiv preprint arXiv:1912.04871* (2019).

37.   Udrescu, S. M. & Tegmark, M. AI Feynman: A physics-inspired method for symbolic regression. *Sci Adv* 6, (2020).

38.   Wolfram, S. The Mathematica Book. *Wolfram Media* (2003).

39. Li, X., Zhang, D. & Li, S. A multi-continuum multiple flow mechanism simulator for unconventional oil and gas recovery. *J Nat Gas Sci Eng* 26, 652–669 (2015).

40. Cao, R., Padilla, E. M. & Callaghan, A. H. The influence of bandwidth on the energetics of intermediate to deep water laboratory breaking waves. *J Fluid Mech* **971**, A11 (2023).

41. Rapp, R. J. & Melville, W. K. Laboratory measurements of deep-water breaking waves. *Philosophical Transactions of the Royal Society of London. Series A, Mathematical and Physical Sciences* **331**, 735–800 (1990).

42. Cao, R., Padilla, E. M., Fang, Y. & Callaghan, A. H. Identification of the free surface for unidirectional nonbreaking water waves from side-view digital images. *IEEE Journal of Oceanic Engineering* (2024).

43. Taylor, P. H., Tang, T., Adcock, T. A. A. & Zang, J. Transformed-FNV: Wave forces on a vertical cylinder — A free-surface formulation. *Coastal Engineering* **189**, (2024).

44. Liu, P. C. Wavelet spectrum analysis and ocean wind waves. in *Wavelet analysis and its applications* vol. 4 151–166 (Elsevier, 1994).

45. Jinshan, X., Jiwei, T. & Enbo, W. The application of wavelet transform to wave breaking. *Acta Mechanica Sinica* **14**, 306–318 (1998).

46. Kim, J. T., Landajuela, M. & Petersen, B. K. Distilling Wikipedia mathematical knowledge into neural network models. *arXiv preprint arXiv:2104.05930*. (2021).

47. Bornschein, J., Visin, F. & Osindero, S. Small data, big decisions: Model selection in the small-data regime. In *International Conference on Machine Learning* 1035–1044 (PMLR, 2020).

48. Kingma, D. P. & Ba, J. L. Adam: A method for stochastic optimization. in *3rd International Conference on Learning Representations, ICLR 2015 - Conference Track Proceedings* (2015).

49. Raissi, M., Perdikaris, P. & Karniadakis, G. E. Physics-informed neural networks: A deep learning framework for solving forward and inverse problems involving nonlinear partial differential equations. *J Comput Phys* **378**, 686–707 (2019).

# Supplementary Information for the Generative Discovery of Partial Differential Equations by Learning from Math Handbooks

Hao Xu[1,2], Yuntian Chen[1,3,*], Rui Cao[4,5], Tianning Tang[6,7], Mengge Du[8], Jian Li[3], Adrian H. Callaghan[5], and Dongxiao Zhang[1,9,*]

[1] Zhejiang Key Laboratory of Industrial Intelligence and Digital Twin, Eastern Institute of Technology, Ningbo, Zhejiang 315200, P. R. China

[2] Department of Electrical Engineering, Tsinghua University, Beijing 100084, P. R. China

[3] Ningbo Institute of Digital Twin, Eastern Institute of Technology, Ningbo, Zhejiang 315200, P. R. China

[4] College of Oceanic and Atmospheric Sciences, Ocean University of China, Qingdao 266100, P. R. China

[5] Department of Civil and Environmental Engineering, Imperial College London, London, SW7 2AZ, United Kingdom

[6] Department of Engineering Science, University of Oxford, Parks Road, Oxford, OX1 3PJ, United Kingdom

[7] Department of Mechanical and Aerospace Engineering, University of Manchester, Manchester, M13 9PL, United Kingdom

[8] College of Engineering, Peking University, Beijing 100871, P. R. China

[9] Institute for Advanced Study, Lingnan University, Tuen Mun, Hong Kong

[*] Corresponding authors

Email address: ychen@eitech.edu.cn (Y. Chen); dzhang@eitech.edu.cn (D. Zhang)

# 1. Supplementary text

## 1.1 Information on the dataset utilized in the experiments

In the discovery of canonical PDEs, eight PDEs from different fields are employed to demonstrate the performance of the proposed generative framework. The information of these PDEs and their respective datasets are provided in Table S1. For evaluating the performance under sparse data and noise, the Allen–Cahn equation is used. Its dataset consists of grid data of 256 spatial observation points in the domain $x \in (-1,1)$ and 201 temporal observation points in the domain $t \in [0,10]$; thus, the data size is 51,456. In the discovery process of the above PDEs, an ANN consisting of five hidden layers, each with 50 neurons, is utilized to construct the surrogate model. 400 population candidates are generated, and the optimization epochs is 5. The learning rate of EqGPT is set to $1 \times 10^{-5}$, and a sparsity regularization coefficient $\alpha_0 = 0.2$ is applied.

For the discovery of PDEs in complex regions, Poisson's equation has been discovered in the dataset on irregular regions. Here, these datasets are introduced in detail. The simple disk region with a radius of $r=1.5$ is illustrated in Fig. 4a. The Dirichlet boundary condition is $u(x,y)=x\sin(xy)$. The solution is obtained from the simulation via *Mathematica*. The data are collected according to polar coordinates, where $r \in [0.001,1.4926]$ with 200 observation points and where $\theta \in [0,2\pi]$ with 201 observation points. Here, coordinates in the Cartesian coordinate system can be obtained by $x=r\cos(\theta)$, and $y=r\sin(\theta)$. Therefore, there are 40,200 data points in the dataset.

For the "smiley face region" investigated in this work, a region of the smiley face is carved out of the disk region with a radius of 2. This region is defined by the implicit region in *Mathematica*. In this region, the Dirichlet boundary condition is $u(x,y)=\sin(x+y)$. The solution is obtained from the simulation using *Mathematica*. The data are collected in a rectangular region in $x \in [-4,4]$ with 250 observations and $y \in [-4,4]$ with 250 observations. Here, the observation data at the empty locations are deleted. Therefore, there are 44,711 valid data points in the dataset.

For the region of "EITech", which consists of discontinuous glyph boundaries, the Dirichlet boundary condition, $u(x,y)=0$, is adopted for each alphabet. This region is defined by the *boundaryDiscretizeGraphics* function in *Mathematica*, and the solution is obtained from the simulation by *Mathematica*. The data are collected in a rectangular region in $x \in [4.3,141]$ with 800 observations and $y \in [19.6,54.6]$ with 200 observations. Here, the observation data at the empty locations are deleted. Therefore, there are 48,367 valid data points in the dataset.

For the 3-dimensional (3D) space shuttle region, the Dirichlet boundary condition is as follows: $u(x,y,z)=1$ if $z \leq 1.3$ and $u(x,y,z)=0$ if $x \leq -7$. This region is defined by the *boundaryDiscretizeGraphics* function in *Mathematica*, and the solution is obtained from the simulation by *Mathematica*. The data are collected in a rectangular region with $x \in [7.65,7.04]$ with 300 observations, $y \in [-4.68,4.68]$ with 200 observations, and $z \in [-1.35,4.16]$ with 100 observations. Here, the observation data at the empty locations are deleted. Therefore, there are 438,960 valid data points in the dataset. In the discovery process, the settings are similar to those described above while the sparsity regularization coefficient $\alpha_0=1$.

For the vibration of an H-shaped elastic membrane, the initial condition is $u(0,x,y)=2e^{-125(x-0.25)^2+(y-0.5)^2}$, and $u_t(0,x,y)=0$. The boundary condition is the Dirichlet boundary, $u(t,x,y)=0$. This region is defined by the *RegionDifference* function in *Mathematica*, which is carved out of the rectangular region where the length of $x$ is 2 and the length of $y$ is 1. The excavated parts are $x \in [0.9,1.1] \cup y \in [0,0.4]$ and $x \in [0.9,1.1] \cup y \in [0.6,1.0]$. The solution is obtained from the simulation using *Mathematica*. The data are collected in a rectangular region in $x \in [0,2]$ with 200 observations, $y \in [0,1]$ with 100 observations, and $t \in [0,2]$ with 200 observations. Here, the observation data at the empty locations are deleted. Therefore, there are 3,774,981 valid data points in the dataset. In the discovery process, the settings are similar to those described above, while 800 population candidates are generated and the sparsity regularization coefficient $\alpha_0=1$ is applied.

For the oil-water two phase flow in 3-dimensional space, the data are generated from the Unconventional Oil and Gas Simulator (UNCONG) under the assumptions of incompressibility and negligible capillary pressure. The simulation domain consists of a three-dimensional discretized geological formation, represented by a structured grid. Two injection wells are positioned at the upper-left and lower-left corners of the grid, while a single production well is located at the center of the right boundary. From the initial day of simulation, water is continuously injected through the injection wells to displace oil within the reservoir and promote its migration toward the production well. The spatial grid is defined over the domain $[x,y] \in [0,30]$ with a uniform spacing of $\Delta x = \Delta y = 1m$, and $z \in [0,8]$ with $\Delta z = 1m$. Temporal discretization spans $t \in [0,30]$ with a time step of $\Delta t = 1$ day. The simulation yields the spatial-temporal distribution of the water-phase pressure field $P_w$, saturation field $S_w$, and effective permeability field $K_w$ and $K_o$, each represented by 600,000 grid-based data points. Finite difference is conducted on the grid-data for differentiation. During PDE discovery, potential differential terms associated with $S_w$, $K_w$, and $K_o$, such as $\partial S_w/\partial t$, $\nabla K_w$, $\nabla K_o$, $\Delta K_w$, and $\Delta K_o$, are incorporated into the vocabulary, while terms related to $P_w$ are constructed analogously to the single-variable setting. The first token is fixed to be $\partial S_w/\partial t$. The symmetry in the $x$ and $y$ directions is used to constrain the generated equations. 600 population candidates are generated, and the optimization epochs is 5. The learning rate of EqGPT is set to $1 \times 10^{-5}$, and a sparsity regularization coefficient $\alpha_0=0.1$ is applied.

For the discovery of previously undisclosed governing equations for highly nonlinear surface gravity waves propagating toward breaking, the data come from real-world wave-tank experiments. A total of 12 independent experiments are conducted with different initial conditions of the wave paddle, which is displayed in Table S3. The spatial surface elevation recorded in each experiment was reconstructed at different time instants before the onset of breaking by employing the image-processing technique. For each experiment, the recorded time period spans between 40 and 100 frames. Each frame contains approximately 4,800 spatial points, covering three non-contiguous spatial subdomains $x \in [8.1516,9.36] \cup [9.74,10.95] \cup [11.392,12.574]$. Data from 12 individual experiments are aggregated for joint PDE discovery. For each experiment, a surrogate model is constructed using an artificial neural network (ANN) comprising six hidden layers with 60 neurons per layer. Each network is trained for up to 50,000 iterations. The input is non-dimensional variable $x^*$ and $t^*$, and the output is the non-dimensional surface elevation $\eta^*$. Automatic differentiation is employed to compute spatial and temporal derivatives from the trained surrogate models. During the PDE discovery process, a population size of 400 candidate equations is used, with five optimization epochs. The learning rate for the EqGPT model is set to $1\times10^{-5}$, and a sparsity regularization coefficient $\alpha_0$=0.02 is applied in the reward function to encourage parsimonious expressions. For each generated PDE, the least squared regression is performed on the data from each of the 12 experiments independently. The individual rewards from these experiments are then averaged to obtain a global reward score, which is used to guide the optimization of the generative model.

## 1.2 Comparison between EqGPT learning from the dataset with and without the target PDE

In this work, to guarantee the validity of the experiments, the target PDE is temporally deleted from the dataset, and the EqGPT generative model is retrained by the modified dataset. This scenario presents a higher level of complexity, closely resembling real-world situations in which the underlying governing equations remain undiscovered. Notably, in practical applications, there is also the possibility that the underlying governing equations align with existing PDE systems, albeit under various conditions. Therefore, the EqGPT trained on the entire PDE dataset can handle practical scenarios. Here, a comparison between the discovery of PDEs with EqGPT trained on the dataset with and without the target PDE is conducted. Similarly, 8 canonical PDEs are investigated. Here, 10 independent experiments with different random seeds are conducted when discovering each PDE, and the success rate is defined as:

$$s_r = n_{success}/n_{total}\,,\tag{S.1}$$

where, $s_r$ refers to the success rate; $n_{success}$ refers to the number of times the correct equation form has been successfully discovered; and $n_{total}$ refers to the number of experiments in total (10 in this work). The results are provided in Fig. S2. The figure shows that the success rate with the target equation (red line) is slightly higher than that without the target equation (blue line). However, even when trained without a target equation, EqGPT achieves success rates above 0.7 in all of the experiments, with minimal to no adverse effects in most instances. This finding demonstrates that the proposed method maintains robust stability under challenging conditions. Notably, it is found that the Chaffee–Infante equation is most affected, suggesting that generating the term $u^3$ may be particularly challenging without learning the target PDE form.

The average discovered epochs with and without the target PDE are also provided in Fig. S2; it is measured by the average of the epochs for which the PDE is first discovered in all 10 independent experiments, which can reflect the optimization efficiency of the method. Notably, for the cases in which the PDE cannot be discovered successfully, the number of epochs is regarded as 5. The figure shows that learning the target form does not strongly affect the optimization efficiency since, in both situations, the average number of discovered epochs is lower than 3, which indicates high speed. Interestingly, the average number of discovered epochs even decreases without learning the target form in some cases, such as the convection diffusion equation and the KdV equation. Similarly, the Chaffee–Infante equation is most affected, which shows an increase of nearly 1 epoch when the target PDE is not learned.

## 1.3 Performance of PDE discovery with different numbers of PDE training EqGPT

In this work, the EqGPT model is trained on the established PDE dataset, which consists of 221 PDEs. To examine the potential of EqGPT, different numbers of PDEs are utilised in the dataset to train the EqGPT model, which is employed to discover the PDEs. Here, the KdV equation, Eq.(6.2.12), and Eq. (8.14.1d) are adopted, where 10,000 data points with no noise are randomly selected to train the surrogate model. Moreover, different numbers of PDEs in the dataset, including 0, 5, 10, 25, 50, 100, 200, and 220 PDEs, are randomly selected to train the EqGPT generative model. Notably, the target PDE has been deleted from the dataset. 10 independent experiments are conducted with different random seeds to calculate the success rate. The results are provided in Fig. S3. The success rate clearly presents an increasing trend with the number of PDEs training the EqGPT. When the generative model is not trained with any PDEs, it is in its initial state and is unable to identify the correct PDE. However, upon training with even a small set of 5 PDEs, EqGPT's success rate improves to 0.3 for the KdV equation. Meanwhile, discovering PDEs with more intricate or irregular terms requires exposure to a broader set of equation forms during training. Notably, as the number of PDEs in the training dataset increases, the success rate also increases, highlighting the potential of the proposed framework; this suggests that a more extensive PDE dataset in the future could be instrumental in tackling more complex problems. Notably, training EqGPT with the current PDE dataset achieves a success rate of over 0.8, indicating a commendable level of performance.

## 1.4 Discovery of PDEs in the compound form

The proposed PDE discovery framework is also capable of discovering PDEs in their compound form. Here, a PDE, called PDE_compound, is investigated, the form of which is written as:

$$u_t - 0.2(uu_x)_x = 0 \,. \tag{S.2}$$

This PDE in a compound form has not been recorded in the scientific handbook and thus does not occur in the PDE dataset, which poses a challenge for PDE discovery. The observation dataset is obtained from the simulation via *Mathematica*. The initial condition is $u(0,x)=\sin(\pi x)$, and the boundary condition is $u(t,0)=u(t,1)=0$. The dataset contains the grid data of 200 spatial observation points in the domain $x \in [0,1]$ and 200 temporal observation points in the domain $t \in [0,1]$; thus, the data size is 40,000. Here, 10,000 data points with no noise are randomly selected to train the surrogate model. The results are provided in Fig. S4. The correct PDE form is discovered in the 3rd optimization epoch, although the accuracy is comparatively low, since this PDE is more complex and rarer. Interestingly, the expanded form of this PDE has also been discovered as the suboptimal structure, which is written as:

$$u_t - 0.2uu_{xx} - 0.2u_x^2 = 0 . \tag{S.3}$$

This means that the proposed PDE discovery framework can discover the PDE in both compound form and expanded form, which shows satisfactory flexibility.

### 1.5 The performance of the proposed method under sparse and noisy data

In this section, the performance of the proposed method is examined by discovering the Allen–Cahn equation under different numbers of data points and different levels of noise. The true equation form is written as:

$$u_t - 0.003u_{xx} - u + u^3 = 0. \tag{S.4}$$

This PDE is comparatively difficult to discover, as it comprises four terms and features a minor coefficient for the diffusion term (Fig. S5a). Similarly, the PDE discovery process is carried out on the meta-data generated by the surrogate model trained from observation data. As illustrated in Fig. S5b and S5c, it is evident that the surrogate model is capable of accurately reconstructing the physical field from highly noisy and sparse data. The settings of these algorithms are detailed in the Materials and Methods.

Fig. S5d presents the coefficient errors when using the Allen–Cahn equation under different numbers of data points and different levels of noise. In this context, the red blocks indicate instances in which the algorithm was unable to identify the correct PDE structure, whereas the blue blocks denote successful discoveries. A deeper blue colour indicates a lower coefficient error. The figure reveals that our proposed generative framework achieves satisfactory accuracy and efficient optimization, typically requiring only 1 to 3 epochs. This efficiency is attributed to the learned knowledge of PDE forms in books, which enables the model to generate proper structures for optimization.

**Supplementary Tables and Figures**

**Table S1. The spatial and temporal domains and data points of the dataset for the canonical PDEs utilized in this work.**

| Equation name | Equation form | Spatial domain and points | Temporal domain and points |
|---|---|---|---|
| KdV equation | $u_t + uu_x + 0.0025u_{xxx} = 0$ | $x \in [-1,1)$<br>$n_x = 512$ | $t \in [0,1]$<br>$n_t = 201$ |
| Burgers' equation | $u_t + uu_x - 0.1u_{xx} = 0$ | $x \in (-8,8)$<br>$n_x = 256$ | $t \in (0,10]$<br>$n_t = 201$ |
| Convection-diffusion equation | $u_t + u_x - 0.25u_{xx} = 0$ | $x \in [0,2]$<br>$n_x = 256$ | $t \in [0,1]$<br>$n_t = 100$ |
| Chaffee-Infante equation | $u_t - u_{xx} + u - u^3 = 0$ | $x \in [0,3]$<br>$n_x = 301$ | $t \in (0,0.5)$<br>$n_t = 200$ |
| Wave equation | $u_{tt} - u_{xx} = 0$ | $x \in [0,\pi]$<br>$n_x = 161$ | $t \in [0,2\pi]$<br>$n_t = 321$ |
| KG equation | $u_{tt} - 0.5u_{xx} + 5u = 0$ | $x \in [-1,1)$<br>$n_x = 201$ | $t \in [0,3]$<br>$n_t = 201$ |
| Eq. (6.2.12)[*] | $0.1u_{xt} + u_t + 0.1u_x = 0$ | $x \in [0,5]$<br>$n_x = 500$ | $t \in [0,10]$<br>$n_t = 500$ |
| Eq. (8.14.1d)[*] | $u_t + u_x/x + 0.25u_{xx} = 0$ | $x \in [1,2]$<br>$n_x = 100$ | $t \in [0,1]$<br>$n_t = 251$ |
| PDE_compound | $u_t - 0.2(uu_x)_x = 0$ | $x \in [0,1]$<br>$n_x = 200$ | $t \in [0,1]$<br>$n_t = 100$ |
| Eq. (7.2.12)[*] | $-0.1u_{xxt} + u_t + 0.1u_x = 0$ | $x \in [0,5]$<br>$n_x = 500$ | $t \in [0,10]$<br>$n_t = 500$ |
| Eq. (6.12.28)[*] | $xu_{tt} - 0.1u_{xx} = 0$ | $x \in [0,5]$<br>$n_x = 500$ | $t \in [0,10]$<br>$n_t = 500$ |
| Fujita–Storm equation | $u_t - 0.05(u_x/u^2)_x = 0$ | $x \in [0,5]$<br>$n_x = 500$ | $t \in [0,5]$<br>$n_t = 500$ |

[*]The equation name corresponds to the index of the math handbook, *Nonlinear Partial Differential Equations for Scientists and Engineers*[26].

**Table S2. The discovered equation of 2-dimensional (2D) Burgers' equation under different noise levels.**

| Noise level | Discovered equation |
|:-----------:|:-------------------:|
| 0% | $u_t + 1.01uu_x + 1.01uu_y - 0.0097\Delta u = 0$ |
| 1% | $u_t + 1.00uu_x + 1.02uu_y - 0.0097\Delta u = 0$ |
| 5% | $u_t + 1.02uu_x + 1.01uu_y - 0.0096\Delta u = 0$ |
| 10% | $u_t + 0.99uu_x + 1.03uu_y - 0.0096\Delta u = 0$ |
| 15% | $u_t + 1.87uu_x + 2.07uu_y - 0.0093\Delta u = 0$ |
| 20% | $u_t + 2.19uu_x + 2.30uu_y - 0.0021\Delta u - 58.7u \cdot u_{tt} = 0$ |

**Table S3. The initial conditions of the wave paddle for conducting 12 independent experiments of wave-breaking.** Here, $G$ is the peak enhancement factor, $T_p$ is the peak wave period of the JONSWAP-type spectra, and $A$ is the total amplitude summed across all underlying wave components. $C_1$, $C_2$, and $C_3$ are the regressed coefficients of the discovered equation.

| Case serial number | G | Tp | A | $C_1$ | $C_2$ | $C_3$ |
|:------------------:|:-:|:--:|:--:|:-----:|:-----:|:-----:|
| 1 | 2 | 1.2 | 80 | 1.461 | $1.936 \times 10^{-3}$ | $-7.38 \times 10^{-5}$ |
| 2 | 2 | 1.2 | 90 | 1.38 | $7.779 \times 10^{-4}$ | $-3.89 \times 10^{-5}$ |
| 3 | 2 | 1.2 | 100 | 1.36 | $1.091 \times 10^{-3}$ | $-4.63 \times 10^{-5}$ |
| 4 | 2 | 1.2 | 105 | 1.426 | $1.875 \times 10^{-3}$ | $-9.12 \times 10^{-5}$ |
| 5 | 2 | 1.3 | 90 | 1.489 | $1.873 \times 10^{-3}$ | $-5.61 \times 10^{-5}$ |
| 6 | 2 | 1.3 | 96 | 1.404 | $1.320 \times 10^{-3}$ | $-4.25 \times 10^{-5}$ |
| 7 | 2 | 1.3 | 105 | 1.469 | $1.529 \times 10^{-3}$ | $-6.40 \times 10^{-5}$ |
| 8 | 2 | 1.3 | 130 | 1.371 | $1.096 \times 10^{-3}$ | $-5.39 \times 10^{-5}$ |
| 9 | 3 | 1.2 | 83 | 1.575 | $2.376 \times 10^{-3}$ | $-7.75 \times 10^{-5}$ |
| 10 | 3 | 1.2 | 95 | 1.511 | $1.967 \times 10^{-3}$ | $-7.45 \times 10^{-5}$ |
| 11 | 3 | 1.2 | 100 | 1.44 | $1.508 \times 10^{-3}$ | $-7.10 \times 10^{-5}$ |
| 12 | 3 | 1.2 | 117 | 1.386 | $4.819 \times 10^{-4}$ | $1.26 \times 10^{-5}$ |

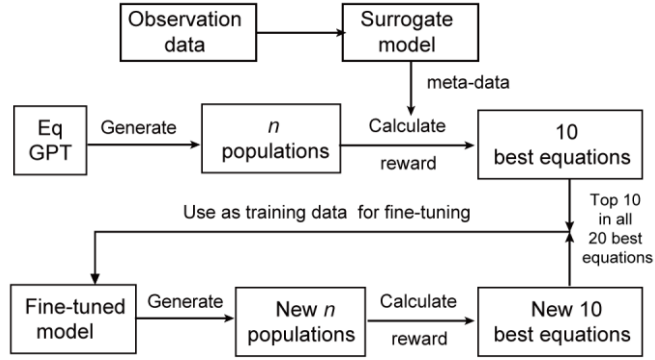

**Fig. S1. Flow chart of the optimization cycle for discovering PDEs.**

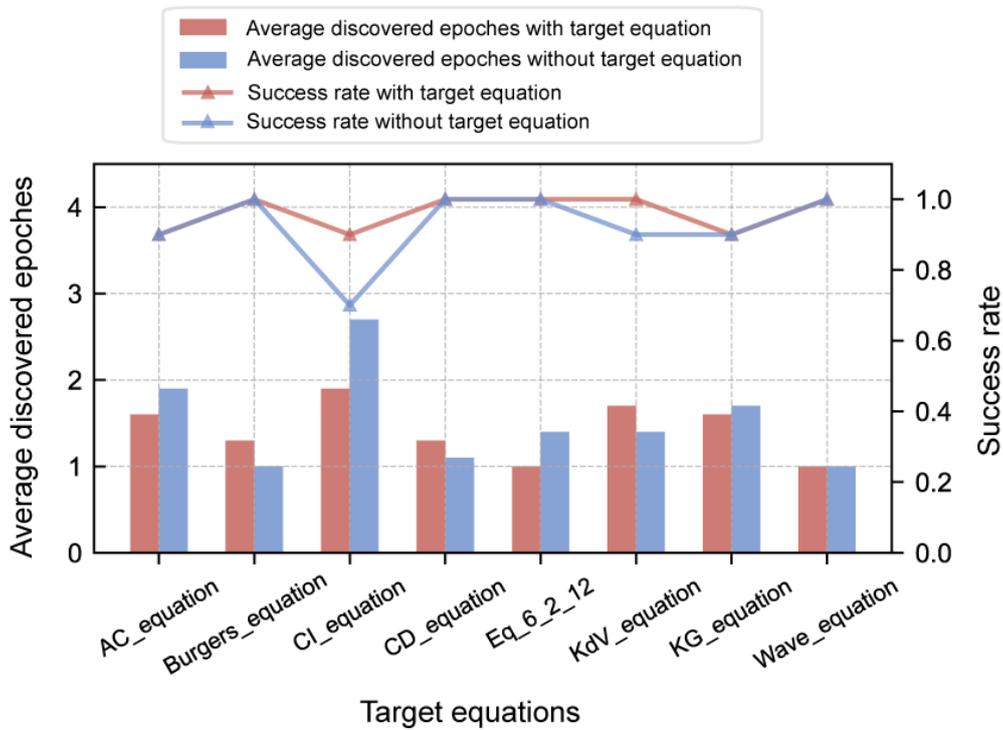

**Fig. S2**. **Comparison between training EqGPT with and without the target equation.** Eight canonical PDEs are investigated, including the Allen Cahn (AC) equation, Burgers' equation, the Chaffee-Infante (CI) equation, the convection diffusion (CD) equation, Eq. (6.2.12) in the handbook, the Korteweg-De Vries (KdV) equation, the Klein–Gordon (KG) equation, and the wave equation. The bar plots show the average number of discovered epochs. The line charts denote the success rate.

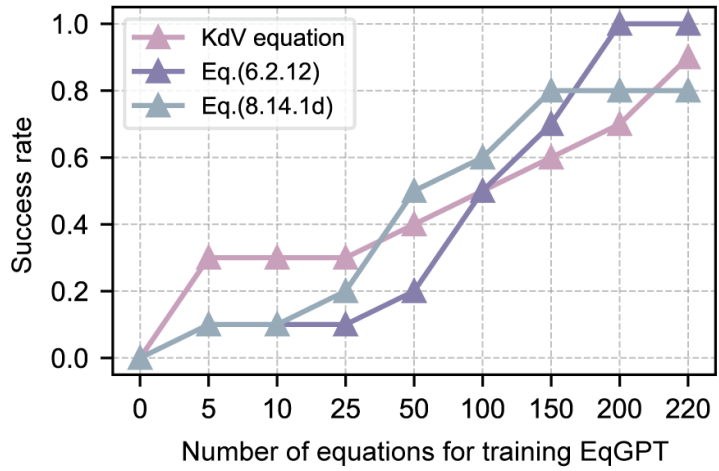

**Fig. S3. Discovery of canonical PDEs with different numbers of equations for training EqGPT.**

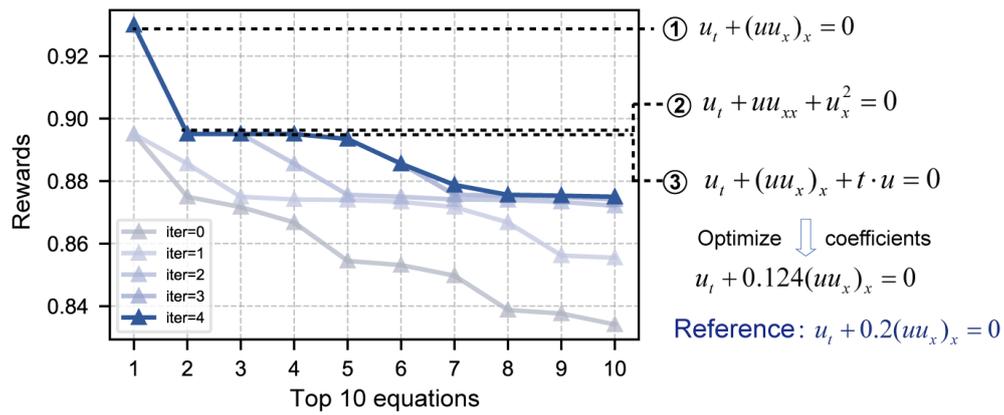

**Fig. S4. Discovery of PDEs in compound form.** The rewards of the top 10 equations generated from the EqGPT model in each optimization epoch **(left)**, and the discovered PDE and the true PDE **(right)**.

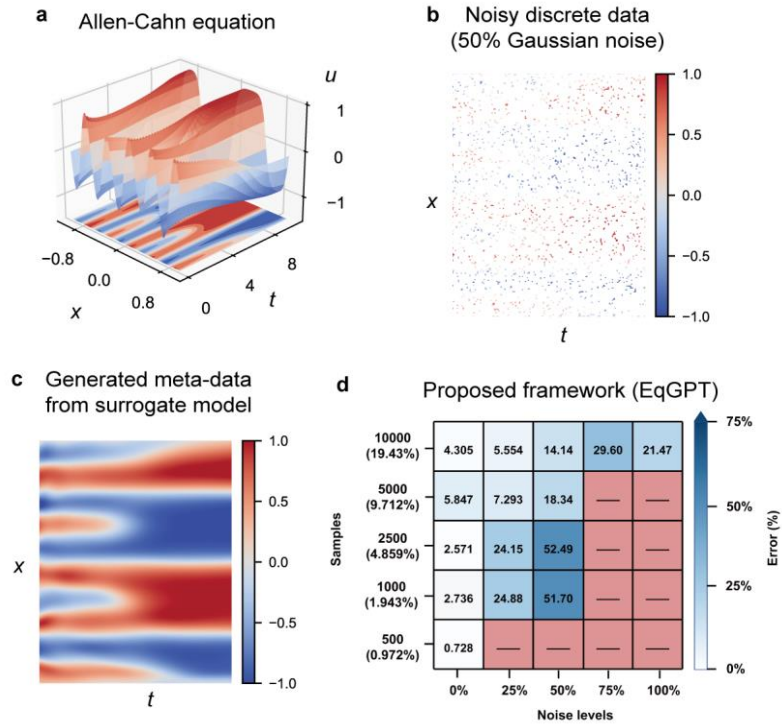

**Fig. S5. The performance of the proposed method under sparse and noisy data when discovering the Allen-Cahn equation. (a)** Illustration of the dataset for the Allen–Cahn equation. **(b)** Visualisation of noisy discrete data with 50% noise. **(c)** The reconstructed metadata from the surrogate model trained on observation data. **(d)** The coefficient error for discovering the Allen–Cahn equation under different levels of noise and data points. The red blocks indicate instances in which the algorithm was unable to identify the correct PDE structure, whereas the blue blocks denote successful discoveries. A deeper blue colour corresponds to a lower coefficient error.